%% file: main.tex
\documentclass[11pt]{article}

\usepackage{acl}

\usepackage{times}
\usepackage{latexsym}
\usepackage[T1]{fontenc}
\usepackage[utf8]{inputenc}
\usepackage{microtype}
\usepackage{inconsolata}

\usepackage{amsmath}
\usepackage{amssymb}
\usepackage{mathtools}
\usepackage{amsthm}

\newtheorem{proposition}{Proposition}

\usepackage{graphicx}
\usepackage{subcaption}
\usepackage{wrapfig}

\usepackage{booktabs}
\usepackage{threeparttable}
\usepackage{multirow}
\usepackage{array}
\usepackage{tabularx}
\usepackage{makecell}
\usepackage{adjustbox}
\usepackage{longtable}

\usepackage{algorithm}
\usepackage{algorithmic}

\usepackage{xcolor}
\usepackage{colortbl}
\usepackage{tcolorbox}
\tcbuselibrary{skins,breakable,listings,theorems}
\definecolor{mygreen}{RGB}{166,207,152}
\definecolor{myblue}{RGB}{235, 245, 255}
\definecolor{output-white}{RGB}{255,255,255}
\definecolor{output-black}{RGB}{0,0,0}
\definecolor{hiddendraw}{RGB}{100, 180, 240}

\providecommand{\corr}{\operatorname{corr}}
\providecommand{\simfunc}{\operatorname{sim}}

\usepackage{url}

\usepackage{enumitem}
\usepackage{listings}
\usepackage{balance}
\usepackage{placeins}
\usepackage{url}
\usepackage{cleveref}

\usepackage{fvextra}

\usepackage{tikz}
\usepackage{forest}
\useforestlibrary{edges}

\title{Task-Utility IQA: Rethinking Image Quality Assessment\\ in Embodied Tasks}

\author{
 \textbf{Jirong Zha\textsuperscript{1}$^*$},
 \textbf{Yanwei Liao\textsuperscript{2}$^*$},
 \textbf{Yemin Wang\textsuperscript{3}$^*$},
 \textbf{Xiangmin Yi\textsuperscript{1}},
\\
 \textbf{Siqi Peng\textsuperscript{4}},
 \textbf{Yingfeng Chen\textsuperscript{1}},
 \textbf{Chen Gao\textsuperscript{1$^\dagger$}},
 \textbf{Xinlei Chen \textsuperscript{1}$^\dagger$}
\\
\\
 \textsuperscript{1}Tsinghua University,
 \textsuperscript{2}Xi'an Jiaotong University,
 \textsuperscript{3}Xiamen University, \\
 \textsuperscript{4}University of Science and Technology of China
\\
 \small{
   \textbf{Correspondence:} \href{chgao96@gmail.com}{chgao96@gmail.com},
   \href{chen.xinlei@sz.tsinghua.edu.cn},
   {chen.xinlei@sz.tsinghua.edu.cn}
 }
 \thanks{These authors contributed equally. $^\dagger$ Corresponding author.}
}

\begin{document}

\maketitle

\begin{abstract}
Image quality assessment (IQA) underpins embodied imaging pipelines by judging whether visual quality satisfies downstream tasks, yet most existing methods learn task-agnostic scores aligned with generic human ratings on static benchmarks. This objective mismatches embodied and interactive settings, where image adequacy depends on task goals, context, and action requirements that shape an agent's decisions. We argue that IQA should shift from score regression to goal-conditioned judgment defined by embodied task utility. To this end, we propose an MLLM-based embodied IQA agent framework that reasons about task requirements, grounds quality evidence, and produces decision-facing assessments. A compact VisDrone evidence study shows that conventional perceptual metrics and generic MLLM-based IQA scores are weak proxies for downstream VLM task success, while a task-conditioned utility score better predicts whether visual input supports the target task. These findings support a task-utility-oriented reframing of IQA for embodied, explainable, and decision-aware visual perception. Code and processed evidence-study results will be released at:
\url{https://github.com/ywliao0829/task-utility-iqa}.
\end{abstract}

\input{sections/Introduction}
\input{sections/Background}
\input{sections/Necessity}

\input{sections/Empirical_Evidence}
\input{sections/Framework}

\input{sections/Implementation}
\input{sections/Alternative_views}
\input{sections/Conclusion}
\input{sections/Impact_statement}


\bibliography{reference}

\newpage

\input{sections/Appendix}




\end{document}

%% file: sections/Introduction.tex
\section{Introduction}
Image quality assessment (IQA) is a foundational problem in computer vision, concerned with determining whether visual signals are suitable for downstream use \cite{zhang2023blind,li2025q,zha2025aircopbench}. IQA underpins a broad range of vision pipelines, including image acquisition and enhancement, compression and streaming, video frame selection, and the evaluation of restoration and generation models \cite{wang2004image,hore2010image,saad2012blind,mittal2012no,ma2025survey}. With the emergence of embodied intelligence, the role of IQA extends beyond perceptual fidelity to embodied perception, where visual observations directly influence perception, reasoning, planning, and action in vision--language--action systems.


\begin{figure}[t]
    \centering
    \includegraphics[width=1\linewidth]{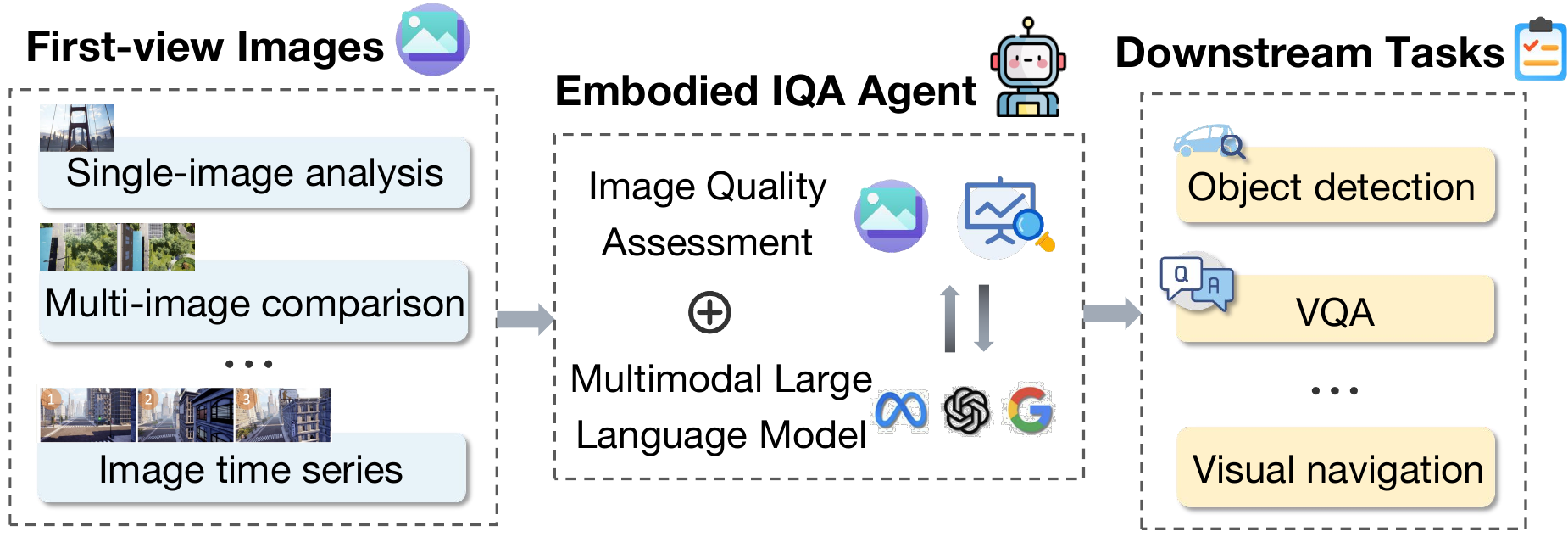}
    \caption{By coupling MLLM reasoning with image quality assessment, the MLLM-based embodied IQA agent evaluates quality from a first-person perspective, incorporating task context, environmental constraints, and goal-driven decision requirements to support autonomous decision-making.}
    \label{fig:overview}
    \vspace{-0.6cm}
\end{figure}

Existing IQA research has progressed from classical full-reference metrics, such as PSNR \cite{hore2010image} and SSIM \cite{wang2004image}, to no-reference methods \cite{saad2012blind,mittal2012no} and deep models optimized on human opinion scores \cite{bosse2017deep,ma2025survey,li2025image}. While these advances have substantially improved perceptual quality prediction, the underlying paradigm remains largely centered on task-agnostic scalar scoring. In embodied settings, however, image quality cannot be fully characterized by perceptual fidelity alone; instead, it should be evaluated according to downstream usability, interpretability, and decision relevance for intelligent agents \cite{zhang2024embodied,zhu2025agenticiqa}. Existing IQA protocols primarily assess alignment with human subjective judgments on static benchmarks, implicitly assuming that perceptual preference correlates with practical utility.


\textbf{Dilemma of Traditional IQA.} In embodied settings, traditional IQA fails when quality requirements vary across tasks, when task-relevant regions occupy only part of the image \cite{chen2024topiq,song2025segmenting}, and when semantic coverage matters more than pixel-level fidelity \cite{kerkouri2025modeling}. In unmanned aerial vehicle (UAV) infrastructure inspection, a task-agnostic metric can assign similar penalties to blur regardless of where it occurs, although blur in sky or background regions is often irrelevant while blur over the inspected surface can make an image unusable. In robotic manipulation, noise or compression may be tolerable for coarse navigation but harmful when edges, keypoints, or depth cues are needed for grasp planning. These cases reveal a fundamental evaluation misalignment: a single quality score optimized for generic human ratings cannot capture diverse embodied task requirements.

\vspace{-0.28cm}
{\setlength{\fboxsep}{6pt}%
\begin{center}
\fcolorbox{black}{gray!10}{\parbox{0.95\linewidth}{%
\textbf{Primary position:} \emph{IQA should embrace MLLM-based embodied agents}, reframing image quality assessment as \textbf{goal-conditioned, explainable, and adaptive judgment} that is evaluated by \textbf{task utility} rather than by correlation with task-agnostic scores.}}
\end{center}}

\textbf{The Turning Point.} IQA stands at an inflection point driven by two converging trends. First, embodied AI agents, including UAVs, mobile robots, and AR systems, create new demands for context-aware quality assessment \cite{morgenthal2014quality,zhang2024embodied,lv2025ar}. Second, multimodal large language models (MLLMs) \cite{liu2023visual} combine visual perception with language-conditioned reasoning, enabling systems that understand context, generate explanations, and adapt to diverse tasks. Recent work shows that MLLMs can express quality judgments in natural language \cite{wu2023q,huang2024visualcritic}, support comparative evaluation \cite{zhu2024adaptive,wu2024towards}, and provide localized rationales \cite{you2024depicting,chen2024q}. The missing step is to integrate IQA into the embodied loop, conditioning judgments and explanations on specific perception tasks and context so they can drive actionable decisions.

\textbf{Why This Position Paper?} As AI systems transition from passive predictors to agents that perceive and act, visual quality should be defined relative to goals, constraints, and actions. Existing IQA objectives and benchmarks do not capture this dependency, and they provide little support for explanation, interaction, and policy-level decisions such as retake, reposition, or proceed. The maturation of MLLMs and the deployment of embodied systems make it timely to articulate a principled research agenda for embodied IQA agents.

\textbf{Contributions.} This position paper makes four contributions: (1) We propose an MLLM-based embodied IQA agent framework that reframes image quality assessment from task-agnostic score regression to goal-conditioned, explainable, and utility-aware judgment; (2) We analyze failure modes of scoring-based IQA in embodied settings and motivate a task-utility view of quality, where visual adequacy depends on task goals, context, and action requirements; (3) We review MLLM-based IQA and recent embodied IQA efforts, highlighting the gap between generic or robot-oriented quality scoring and downstream task-utility alignment; (4) We provide compact empirical evidence on VisDrone~\citep{zhu2018vision,du2019visdroneDet} showing that task-agnostic perceptual and generic MLLM-based IQA scores are weak proxies for downstream task success, while a task-conditioned utility score better aligns with task utility. We provide an extended literature taxonomy and supplementary formulation details in Appendix~\ref{app:literature_review} and Appendix~\ref{app:training_formulations}.

\begin{figure*}[!t]
    \centering
    \includegraphics[width=0.8\linewidth]{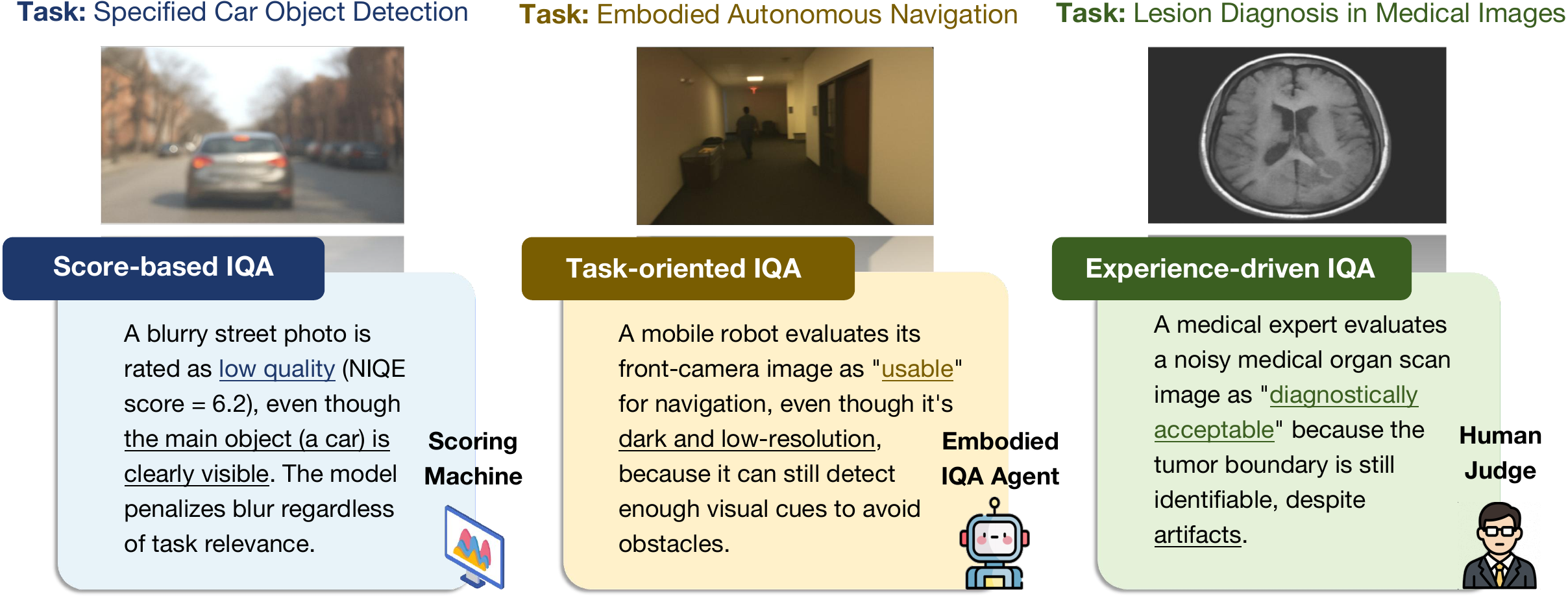}
    \caption{Illustration of low-quality yet high-utility image cases for different downstream tasks, showing how score-based IQA, embodied IQA agents, and human experts prioritize different criteria. While traditional IQA reduces quality to a single score, embodied agents and human experts focus on task-relevant regions and utility. This contrast highlights the need to rethink IQA as goal-conditioned, task-aware judgment rather than task-agnostic scoring.}
    \label{fig:background}
\end{figure*}

\input{tables/comparison}

%% file: tables/comparison.tex
\begin{table*}[!t]
\centering
\caption{Capability comparison of existing IQA approaches with embodied IQA agents.}
\label{tab:comparison}
\begin{threeparttable}
\small
\begin{tabular}{lccc}
\toprule
\textbf{Capability} & \textbf{Traditional IQA Scoring} & \textbf{MLLM-based IQA Reasoning} & \textbf{Embodied IQA Agent} \\
\midrule
Numerical assessment & \checkmark & \checkmark & \checkmark \\
Task awareness & $\times$ & $\sim$ (if prompted) & \checkmark \\
Explanation & $\times$ & \checkmark & \checkmark \\
Task-grounded explanation & $\times$ & $\times$ & \checkmark \\
Context adaptation & $\times$ & $\sim$ (if fine-tuned) & \checkmark \\
Embodied integration & $\times$ & $\times$ & \checkmark \\
\bottomrule
\end{tabular}
\begin{tablenotes}
\small
\item \textbf{Note:} \checkmark indicates full support; $\times$ indicates lack of support; $\sim$ indicates partial support.
\end{tablenotes}
\end{threeparttable}
\end{table*}

%% file: sections/Background.tex
\section{Background: MLLM-based IQA and Related Work}

Recent work has begun to recast IQA using multimodal foundation models, offering richer outputs than scalar scores. We summarize this line of work along three themes: (1) single-image assessment, (2) comparative and relational evaluation, and (3) grounded quality analysis. While these directions improve interpretability and flexibility, they still rarely model the \emph{task-conditioned} nature of quality required by embodied agents. \autoref{tab:comparison}\footnote{Detailed concept definitions are provided in \autoref{key_concepts}.} highlights the gap between current MLLM-based IQA and embodied requirements.

\subsection{Single-Image IQA with MLLMs}

Early work leveraged pre-trained vision-language models (VLMs) for IQA, establishing foundations for MLLM-based quality assessment. CLIP-IQA \cite{wang2023exploring} demonstrates that CLIP embeddings can assess quality without task-specific training, achieving reasonable performance on general quality assessment. This zero-shot approach reveals that VLMs capture quality-relevant features.

Q-Instruct \cite{wu2024q} moves beyond zero-shot prompting by constructing instruction-tuning datasets that teach MLLMs to perform low-level visual quality judgments through natural language reasoning. By aligning quality assessment with instruction-following capabilities, it demonstrates that instruction tuning substantially improves MLLM performance on IQA-related tasks.

Q-ALIGN \cite{wu2023q} further advances this line of work by replacing continuous quality scores with discrete, text-defined quality levels that better reflect subjective evaluation categories such as \emph{excellent}, \emph{good}, and \emph{fair}. Framing IQA as a classification problem rather than regression, Q-ALIGN achieves state-of-the-art results on KonIQ-10K \cite{lin2018koniq}.
Other recent work includes PromptIQA \cite{chen2024promptiqa}, which uses prompts to boost performance and generalization, and multi-modal prompt learning approaches \cite{pan2024multi} that enhance blind image quality assessment. Teaching-based methods \cite{zhang2025teaching,you2025teaching} explore how to train LMMs for image quality scoring and interpreting. Next Token approaches \cite{li2025next} demonstrate realistic image quality and aesthetic scoring with MLLMs. Align-IQA \cite{yang2024align} addresses diverse human preferences through customizable guidance, enabling more flexible quality assessment aligned with different user needs. A comprehensive study \cite{wu2024comprehensive} systematically investigates psychophysical experimental designs (single-stimulus, double-stimulus, and multiple-stimulus methods) combined with prompting strategies (standard, in-context, and chain-of-thought) for MLLM-based IQA, revealing that while GPT-4V provides reasonable quality assessment, it struggles with fine-grained quality discrimination and multi-image comparison tasks.

Despite these advances, these methods treat IQA as an isolated task without considering task context. CLIP-IQA provides no task-specific reasoning, Q-Instruct focuses on general quality improvements, and Q-ALIGN, while more interpretable, still produces task-agnostic assessments. They do not integrate with embodied decision-making pipelines where quality judgments should inform actions and adapt to changing mission requirements. The gap between general quality assessment and task-aware judgment represents a fundamental challenge that these methods do not address.

\subsection{Comparative and Relational IQA}

Beyond single-image IQA, comparative and relational methods align quality assessment with human judgment by enabling relative, context-aware reasoning across multiple images.
Compare2Score \cite{zhu2024adaptive} uses adaptive soft comparison to estimate pairwise preferences. Q-Insight \cite{li2025q} employs group relative policy optimization for ranked image sets. Co-Instruct \cite{wu2024towards} enables open-ended visual quality comparison and introduces MICBench, a benchmark for multi-image comparison tasks.

The advances in comparative assessment enable richer judgments than scalar scores but remain detached from task context. Pairwise comparison is more informative than assigning a single score, yet task-conditioned comparison is more meaningful for embodied applications; for example, an image should be judged by whether it supports obstacle avoidance rather than by generic visual preference alone. Bridging this gap between general and task-aware comparison is a key opportunity for embodied IQA.

\subsection{Grounded Quality Assessment}

Recent quality assessment work incorporates spatial grounding, recognizing that quality issues are often localized. Q-Ground \cite{chen2024q} detects and describes local distortions via multi-scale feature learning and introduces the QGround-100K dataset, while DepictQA \cite{you2024depicting} generates descriptive quality comparisons across image pairs.

However, these explanations are not task-grounded: they identify what is wrong, but not why it matters for a specific task. Q-Ground can localize blur but cannot assess its task relevance (e.g., structural surfaces vs. sky for crack detection). DepictQA provides descriptive comparisons but cannot relate them to task requirements. As a result, they offer limited actionable guidance for embodied systems where explanations need to inform decisions.

\textbf{Embodied IQA.}
Recent work has begun to explicitly connect IQA with embodied AI. \citet{zhang2024embodied} propose an embodied image quality assessment setting for robotic intelligence, with an embodied preference database and a dedicated no-reference IQA model. More recently, \citet{li2026embodiediqa} formulate IQA for embodied AI as a benchmark problem and construct a larger Embodied-IQA database with fine-grained annotations from VLMs, VLA models, and real-world robots. These works are directly related to our position, and we do not treat them as peripheral baselines. Our focus is complementary but distinct: rather than constructing a robot-oriented quality database or training a dedicated NR-IQA predictor, we ask whether a quality signal can predict downstream task success under a given task and context. This shifts the target from embodied quality scoring to task-utility alignment, where the key question is whether the visual input supports the intended decision.
\vspace{-0.25cm}
{\setlength{\fboxsep}{6pt}%
\begin{center}
\fcolorbox{black}{gray!10}{\parbox{0.95\linewidth}{%
\textbf{Extended position:} Our position extends the advances of MLLM-based IQA by calling for \textit{task-aware grounding}: explanations should not just describe quality degradations but relate them to task utility. The transition from \emph{describing} motion blur in a region to \emph{explaining} how motion blur obscures the target structure needed for navigation represents a fundamental shift.}}
\end{center}}


%% file: sections/Necessity.tex
 \section{The Necessity of Embodied IQA}

\subsection{Empirical Motivation: Typical Embodied Scenarios}
\label{sec:scenarios}
We highlight concrete settings where visual quality is inherently conditional and a task-agnostic score can be misleading. Each illustrates a different reason that embodied systems need goal-conditioned judgment rather than a single global score, supported by real-world applications, empirical evidence, and references to actual deployment scenarios\footnote{The presented examples are grounded in existing literature \cite{morgenthal2014quality,zhang2024embodied,lv2025ar,nguyen2024mcd,abedin2020data} and illustrate why task-agnostic scoring fails in embodied settings.}.

\textbf{Robotic perception and manipulation.} In robot manipulation and navigation, camera inputs are consumed by downstream modules (e.g., pose estimation, tracking, or grasp planning). Quality is therefore tied to whether task-relevant cues (edges, keypoints, depth cues) are preserved under the current objective \cite{zhang2024embodied}. Empirical studies on embodied IQA \cite{zhang2024embodied} suggest that noise or compression may be tolerable for coarse navigation but harmful for fine manipulation tasks requiring precise geometric information. Large-scale robot perception datasets further emphasize the diversity of real-world conditions (lighting, motion, sensor artifacts) that induce such task- and context-dependent requirements \cite{nguyen2024mcd}. Conversely, visually clean imagery can still be inadequate if lighting or viewpoint obscures critical geometry needed for object recognition or pose estimation. For instance, in robotic grasping, an image with high overall quality but poor lighting on the target object's edges may fail to support reliable grasp planning, while a slightly noisier image with adequate edge visibility may enable successful manipulation. This demonstrates how task-specific requirements (edge detection for grasping) override general quality metrics.

\textbf{UAV infrastructure inspection.} In UAV-based structural inspection, operators care about the visibility of task-relevant surfaces and defects, and generic no-reference IQA scores can correlate weakly with inspection outcomes because they do not encode coverage or task-relevant defect visibility \cite{morgenthal2014quality}. UAV deployments also impose constraints on energy and data freshness \cite{abedin2020data}, making quality judgments inherently multi-objective: mild blur can be acceptable when the inspected region remains resolvable, whereas a sharp image that misses the target surface is unusable. A useful IQA judgment should incorporate \emph{coverage} (is the relevant structure present?) and \emph{localized adequacy} (are critical regions sufficiently clear?). For example, during bridge inspection, an image with slight motion blur in non-critical regions (sky, background) may score lower on traditional metrics yet remain fully adequate for crack detection if the structural surface is clearly visible and in focus. This indicates that task-relevant regions (structural surfaces) require different quality criteria than background regions.

\textbf{AR-guided assembly.} In augmented reality (AR) assistance, visual quality should balance detail, contrast, and latency under real-time constraints, and the relevant criterion depends on whether the system needs to recognize parts, align overlays, or support user verification \cite{lv2025ar}. A task-agnostic score does not capture these multi-objective trade-offs or the need for actionable feedback (e.g., increase exposure, adjust viewpoint). For instance, in manufacturing assembly tasks, AR systems should overlay virtual instructions onto physical components; quality assessment is expected to prioritize alignment accuracy and contrast in regions where overlays appear, rather than applying uniform quality criteria across the entire image. Research in AR-assisted assembly \cite{lv2025ar} highlights the importance of task-aware visual assessment for effective overlay alignment, though the specific relationship between quality metrics and alignment precision requires further empirical investigation.

\textbf{Summary:} Across these scenarios, the bottleneck is not merely scoring accuracy but \emph{conditional adequacy}: quality depends on task, context, and action, and effective systems should explain what limits utility and what can be done next. The examples illustrate why traditional scoring-based IQA fails when quality requirements vary across tasks, when semantic coverage matters more than pixel fidelity, and when multi-objective trade-offs should be balanced under real-time constraints.

\begin{figure*}[t]
    \centering
    \includegraphics[width=0.8\linewidth]{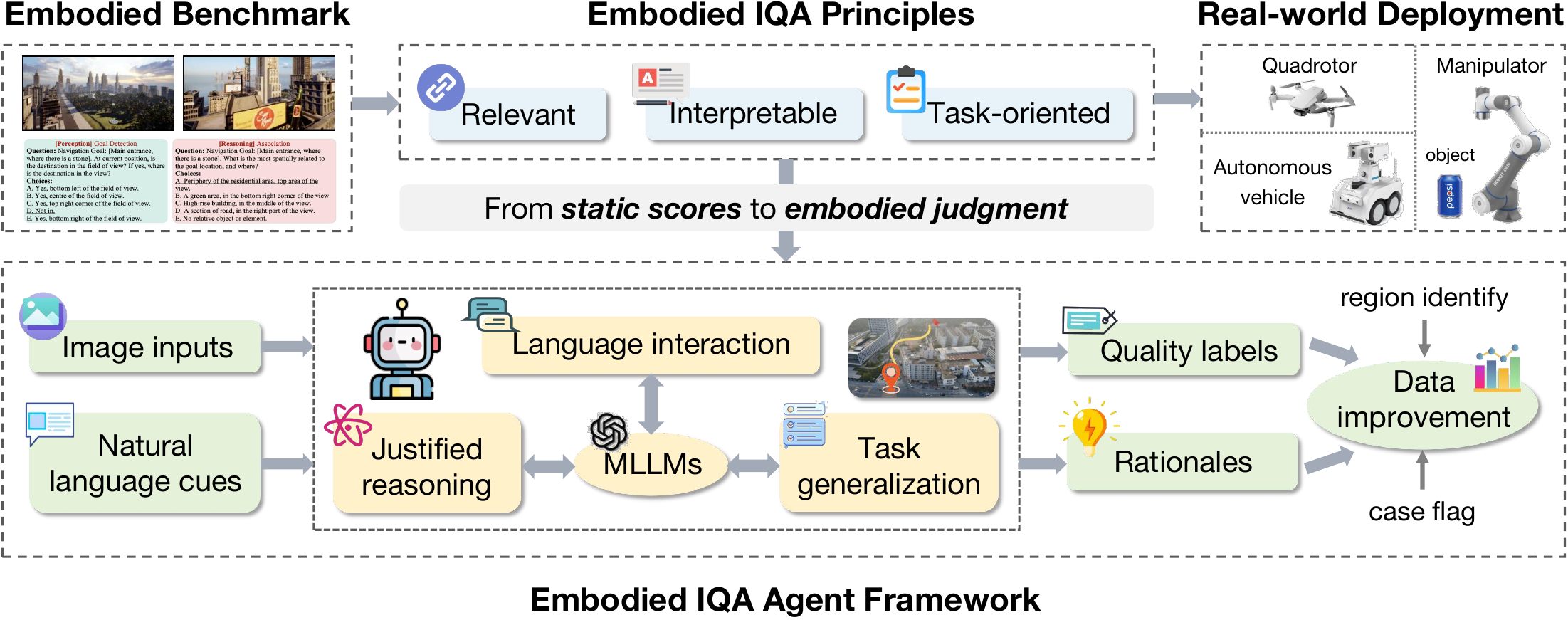}
    \caption{Multi-step decision pipeline of MLLM-based embodied IQA agents: from task understanding to knowledge grounding, visual rationale, tool refinement, and context-aware decision. The framework reconceptualizes IQA as a task-driven, interpretable, and context-relevant reasoning process rather than an absolute perceptual scoring problem.}
    \label{fig:paradigm}
\end{figure*}

Additional UAV inspection analysis and task-specific failure modes are provided in Appendix~\ref{app:uav_case_study}.

\subsection{Theoretical Foundation: A Task-utility Conditional View of Quality}
\label{sec:theory}
We formalize the distinction between traditional and embodied IQA through a conditional framework. Let $I$ denote an image. Traditional IQA typically learns a task-agnostic predictor $Q(I)$ from human opinion scores $H(I)$ by minimizing a loss $\mathbb{E}_{I \sim \mathcal{D}}[\ell(Q(I), H(I))]$, and reports rank/linear correlation (e.g., the Spearman rank correlation coefficient (SRCC) and Pearson linear correlation coefficient (PLCC)) on static test sets \cite{wang2004image,hore2010image,saad2012blind,mittal2012no}. Embodied IQA instead requires a \emph{conditional} judgment $Q(I \mid T, C)$, where $T$ denotes a task specification and $C$ denotes context (environment, constraints, goals). The objective is to maximize expected \emph{task utility}:
\begin{equation}
\mathbb{E}_{(I,T,C) \sim \mathcal{D}}\!\left[U\!\left(Q(I \mid T,C),\, \text{task\_success}(I,T)\right)\right],
\label{eq:utility}
\end{equation}
where $U(\cdot)$ measures how well the assessment supports downstream performance and decision-making.


\begin{proposition}
When task-relevant features differ across tasks, access to $(T,C)$ can yield higher expected task utility than an unconditional assessment $Q(I)$ trained only to match task-agnostic ratings.
\end{proposition}
\vspace{-0.03cm}
Intuitively, a single unconditional predictor must compromise between conflicting task requirements, whereas $Q(I\mid T,C)$ can align its judgment with the task-specific utility signal. This conditional view also supports transfer across related tasks while retaining task-specific decision rules.

%% file: sections/Empirical_Evidence.tex
\section{Empirical Evidence: Task Utility Beyond Perceptual Quality}
\label{sec:empirical_evidence}

To support our position, we conduct a compact evidence study that evaluates image quality by its ability to predict downstream task success rather than by task-agnostic perceptual fidelity. The goal is not to establish a saturated benchmark, but to test whether task-utility alignment exposes a measurable mismatch between conventional IQA scores and embodied visual decision-making. We use VisDrone2019-DET~\citep{zhu2018vision,du2019visdroneDet} as a UAV-view embodied setting and convert detection annotations into VQA-style queries, including positive existence, negative existence, and counting tasks.

Our setting differs from recent embodied IQA benchmarks and models~\citep{zhang2024embodied,li2026embodiediqa}. Existing work mainly studies robot-oriented quality scoring, embodied preference databases, or dedicated NR-IQA models. In contrast, we evaluate whether each IQA or utility score predicts downstream task success under fixed image--question pairs, gold answers, and answer normalization.

\begin{figure*}[!t]
    \centering
    \includegraphics[width=0.92\textwidth]{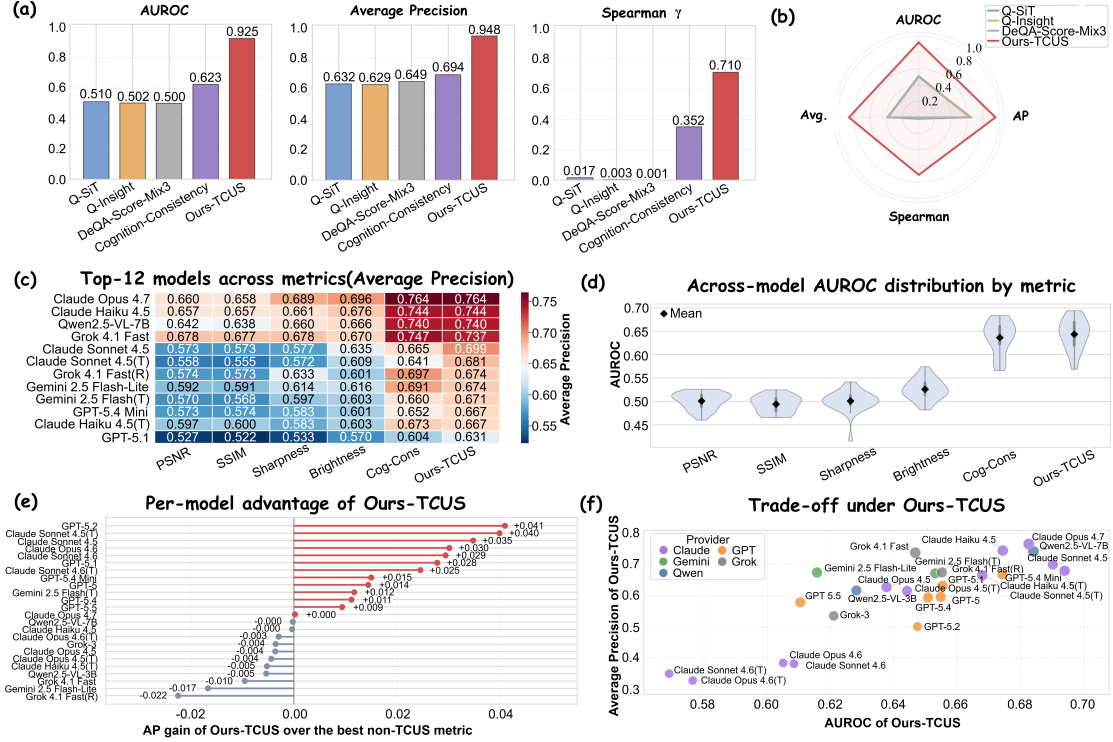}
    \vspace{-0.7em}
    \caption{Task-utility evidence on VisDrone. Each IQA or utility score is evaluated by its ability to predict downstream VLM task success. Panel (a) compares IQA baselines, cognition-consistency, and Ours-TCUS. Panel (b) summarizes raw baseline-level values. Panels (c)--(f) provide cross-model evidence through AP heatmaps, AUROC distributions, per-model gains, and AUROC--AP trade-offs.}
    \label{fig:task_utility_composite_main}
    \vspace{-0.9em}
\end{figure*}

\begin{table}[t]
\centering
\small
\setlength{\tabcolsep}{2.6pt}
\caption{Protocol for measuring task-utility alignment.}
\begin{tabularx}{\columnwidth}{lX}
\toprule
\textbf{Item} & \textbf{Setting} \\
\midrule
Dataset & VisDrone2019-DET UAV-view images. \\
Subset & Unified VisDrone-300 image--question pairs. \\
Tasks & Positive existence, negative existence, and counting. \\
Target VLMs & 16 models from Qwen, GPT, Gemini, Claude, and Grok. \\
Target label & Binary task success after answer normalization. \\
Indicators & PSNR, SSIM, sharpness, brightness, cognition-consistency, and Ours-TCUS. \\
Metrics & AUROC, AP, and Spearman correlation on identical samples. \\
\bottomrule
\end{tabularx}
\label{tab:empirical_protocol}
\vspace{-0.8em}
\end{table}

Table~\ref{tab:empirical_protocol} summarizes the evaluation protocol. For each image--question pair, we first record whether a target VLM answer matches the gold task outcome after normalization, yielding a binary task-success label. We then evaluate whether different quality indicators can predict this label. Ours-TCUS denotes the task-conditioned utility score in our evidence study, combining perceptual evidence with degradation, question-type, and prediction-consistency signals. AUROC and average precision (AP) measure how well a scalar indicator ranks successful task instances above failed ones, while Spearman correlation measures its monotonic association with task success. Traditional metrics are not used to filter samples, alter images, or change VLM inputs; they are evaluated only as predictors of downstream success.

As shown in Figure~\ref{fig:task_utility_composite_main}, generic IQA scores are weak proxies for embodied task success. Q-SiT, Q-Insight, and DeQA-Score-Mix3 remain close to chance-level AUROC, while cognition-consistency improves alignment but still falls short of a task-conditioned utility signal. Ours-TCUS achieves AUROC=0.925, AP=0.948, and Spearman $r=0.710$, with consistent advantage across multiple target VLM families. This evidence supports our central distinction from prior IQA work: embodied IQA should not only estimate perceived or robot-preferred quality, but also measure whether visual input is useful for a specified downstream task. Additional external baselines, qualitative VisDrone cases, and extended evaluation details are provided in Appendix~\ref{app:visdrone_empirical_evidence}.

%% file: sections/Framework.tex
\section{Framework for Embodied IQA Agent}

This section formalizes the paradigm in \autoref{fig:paradigm} as a structured decision pipeline for an MLLM-based embodied IQA agent, where image quality assessment is no longer treated as a static scoring problem but as a goal-driven, knowledge-augmented, and tool-enabled reasoning process coordinated by the MLLM agent. 
Detailed training objectives, algorithmic formulations, and implementation roadmap are provided in Appendix~\ref{app:training_formulations} and Appendix~\ref{app:implementation_roadmap}.

\textbf{Step 1: Task understanding, decomposition, and knowledge grounding.}
Given image inputs and natural language cues of the downstream goal, the MLLM-based embodied IQA agent first interprets the task and decomposes it into sequential subtasks, forming intermediate decision points for quality assessment. To establish appropriate evaluation criteria, the agent retrieves domain-specific expert knowledge through retrieval-augmented generation (RAG), grounding its reasoning in task-relevant standards rather than generic perceptual heuristics.

\textbf{Step 2: Rationale-guided quality judgment with visual grounding.}
Conditioned on the task requirements and retrieved knowledge, the MLLM agent produces quality judgments such as rankings or usability decisions together with explicit textual rationales. These rationales are visually grounded by identifying task-relevant regions in the image, enabling region-level analysis that explains how specific quality factors influence task performance.

\textbf{Step 3: Tool-assisted refinement and utility-aware decision making.}
When quality deficiencies are detected, the MLLM agent can invoke external tools to enhance the image for the current task, such as denoising, contrast adjustment, or cropping task-relevant areas. Throughout this process, the agent estimates its confidence and flags ambiguous or borderline cases. By integrating perception, language-conditioned reasoning, knowledge retrieval, and tool usage, the MLLM-based IQA agent ultimately produces utility-aware quality decisions along with actionable feedback for improving image acquisition or processing parameters.

%% file: sections/Implementation.tex
\section{Implementation of Embodied IQA Agent}

Building embodied IQA agents requires coordinated design across four components: embodied datasets, instruction-based training, grounded evaluation, and edge device deployment. Each component should support task-aware, interpretable, and context-grounded assessment. 
%
Specifically, embodied data defines the task contexts and utility signals that ground task usability; training enables models to internalize these signals through goal-conditioned objectives and shared representations; evaluation measures success by downstream task support rather than score correlation; and deployment situates judgments within real pipelines. Together, these components form a closed loop that links task demands to practical, context-aware quality assessment in embodied systems.


\subsection{Embodied Datasets}
Current IQA datasets \cite{lin2018koniq} focus on general quality or specific distortions. Embodied IQA requires datasets that pair images with task specifications, environmental context, and goal-driven annotations. Recent embodied datasets \cite{wang2024embodiedscan,yao2024aeroverse,gao2024embodiedcity} provide visual data from embodied platforms but lack quality annotations. Extending these with task-aware quality labels would support embodied IQA development. The annotation schema should include task specifications, quality judgments, explanations, and task performance outcomes.

\subsection{Instruction-based Training}
MLLMs should be trained to follow task-specific prompts, such as \emph{assess image quality for obstacle detection}, and produce reasoned outputs. Instruction tuning \cite{wu2024q} enables this capability, but current approaches focus on general quality rather than task-specific judgment. Training should incorporate pairwise comparisons, soft-label regression, and human feedback to align model judgments with task utility.

Direct Preference Optimization (DPO) \cite{rafailov2023direct} and human feedback techniques \cite{christiano2017deep} can align IQA judgments with task performance rather than abstract quality scores. The training objective can combine utility, explanation, and consistency losses; detailed formulations are provided in Appendix~\ref{alg:training}, Appendix~\ref{eq:loss}, and Appendix~\ref{eq:dpo}.

\subsection{Grounded Evaluation and Deployment}
\label{sec:eval}
Standard IQA correlations such as PLCC and SRCC measure agreement with human ratings but not task utility. Embodied IQA should evaluate usefulness, explanation fidelity, and cross-context consistency, while deployment requires lightweight real-time systems integrated with decision-making pipelines. Distillation \cite{hinton2015distilling,zhang2024embodied} and quantization \cite{rastegari2016xnor} can support edge devices; detailed protocols are provided in Appendix~\ref{sec:appendix_eval}.

%% file: sections/Alternative_views.tex
\section{Alternative Views}
\vspace{-0.1cm} 
A strong counterargument is that embodied IQA may not require MLLM-based agents. One could select task-specific IQA metrics or train lightweight predictors on embodied IQA datasets, which is attractive because MLLMs are expensive, can hallucinate, and are difficult to deploy in real time. Such cheaper predictors may be preferable when task labels are available and explanations are unnecessary.

We agree that MLLM-based agents should not replace all IQA metrics. In low-risk or well-specified settings, cheap perceptual metrics or lightweight predictors may be preferable. Our claim is narrower: when quality depends on task semantics, contextual constraints, target regions, and actionable explanations, scalar perceptual scores are insufficient. As Table~\ref{tab:comparison} shows, MLLM-based IQA reasoning provides explanations and partial task awareness when prompted, but does not ensure task-grounded explanation, context adaptation, or embodied integration. An embodied IQA agent adds this task-grounding and decision-facing layer.

A second alternative is to optimize machine-specific perception rather than human-like judgment. We view this as a design trade-off: embodied IQA should preserve human-aligned interpretability for safety and oversight, while exploiting machine consistency where scale or latency matters. The key question is whether the judgment supports the task under the current context; deployment risks are discussed in Appendix~\ref{app:practical_risks}.

%% file: sections/Conclusion.tex
\section{Conclusion}
This position paper argues that IQA for embodied settings should be evaluated by task utility rather than task-agnostic perceptual fidelity alone. The key question is whether an image preserves the information needed for a specific downstream decision. We advocate a shift from generic quality scoring to goal-conditioned usability assessment, connecting IQA with embodied AI, multimodal reasoning, and safety-aware decision making. Our VisDrone evidence shows that conventional perceptual metrics and recent MLLM-based IQA baselines are weak predictors of downstream task success, while task-conditioned utility signals better align with embodied visual usefulness. This framing motivates benchmarks, models, and evaluation protocols for task-grounded embodied IQA.

%% file: sections/Impact_statement.tex
\section*{Limitations}

This paper is a position paper supported by a compact VisDrone evidence study rather than a saturated benchmark. The current evidence focuses on UAV-view VQA-style perception tasks derived from detection annotations, so it does not yet cover the full range of embodied settings, closed-loop policies, or human-agent decision scenarios. Ours-TCUS is also an empirical proxy for task-utility alignment rather than a final metric, and its generality should be further validated across broader tasks, sensors, and deployment conditions.

Current MLLMs may hallucinate, misground rationales, overlook fine-grained visual evidence, or produce unstable judgments under distribution shift. Their computational cost can also be prohibitive for real-time embodied systems. Future work will extend the evaluation to broader embodied domains, collect richer task-performance labels, develop uncertainty-aware and lightweight utility predictors, and validate task-utility IQA in closed-loop decision-making settings.

\section*{Ethical Considerations}

Task-utility-oriented IQA may improve the reliability of embodied systems by prioritizing whether visual inputs support safe downstream decisions. However, task-conditioned judgments may be over-trusted in safety-critical settings such as UAV inspection, autonomous driving, or robotic operation. Responsible deployment requires uncertainty reporting, domain-specific validation, and human oversight under distribution shift.

%% file: sections/Appendix.tex
\appendix
\onecolumn
\section{Literature Review}
\label{app:literature_review}
\input{figures/taxonomy_v3}

\section{Training Formulations and Algorithms}
\label{app:training_formulations}

\subsection{Training Objective}

The training objective for embodied IQA can be formalized as follows. Given a dataset $\mathcal{D} = \{(I_i, T_i, C_i, P_i, E_i)\}$ where $I_i$ is an image, $T_i$ is a task, $C_i$ is context, $P_i$ is task performance outcome, and $E_i$ is an expert explanation, we optimize:

\begin{equation}
\mathcal{L} = \lambda_1 \mathcal{L}_{\text{utility}} + \lambda_2 \mathcal{L}_{\text{explanation}} + \lambda_3 \mathcal{L}_{\text{consistency}},   
\label{eq:loss}
\end{equation}

where $\mathcal{L}_{\text{utility}} = -\mathbb{E}[\corr(Q(I|T,C), P)]$ measures correlation between quality assessment and task performance, $\mathcal{L}_{\text{explanation}} = \simfunc(E_{\text{model}}, E)$ measures explanation similarity to expert annotations, and $\mathcal{L}_{\text{consistency}}$ enforces that similar tasks produce similar quality criteria while different tasks allow adaptation.

\subsection{Direct Preference Optimization}

For preference-based training, we use pairs $(I_1, I_2)$ where preference is determined by task performance: if task success is higher with $I_1$ than $I_2$, the model should prefer $I_1$ even if $I_2$ scores higher on traditional metrics. Let $x_i=(I_i,T,C)$ denote the conditioning input. The DPO objective becomes:


\begin{equation}
\mathcal{L}_{\text{DPO}} = -\log \sigma \left( \beta \left[ \log \frac{\pi_\theta(y_1 \mid x_1)}{\pi_\theta(y_2 \mid x_2)} - \log \frac{\pi_{\text{ref}}(y_1 \mid x_1)}{\pi_{\text{ref}}(y_2 \mid x_2)} \right] \right),
\label{eq:dpo}
\end{equation}

where $y_1, y_2$ are quality judgments, $\pi_\theta$ is the policy being optimized, $\pi_{\text{ref}}$ is a reference policy, and $\beta$ controls the strength of the preference signal.

\subsection{Training Algorithm}

\begin{algorithm}[t]
\caption{Embodied IQA Training with Task-Utility Alignment}
\label{alg:training}
\begin{algorithmic}[1]
\REQUIRE Dataset $\mathcal{D} = \{(I_i, T_i, C_i, P_i, E_i)\}$ where $P_i$ is task performance
\REQUIRE Pre-trained MLLM $f_\theta$, learning rate $\alpha$, DPO coefficient $\beta$
\ENSURE Fine-tuned model $f_{\theta^*}$
\STATE Initialize $\theta$ from pre-trained MLLM weights
\FOR{epoch $= 1$ to $E$}
    \FOR{batch $(I_b, T_b, C_b, P_b, E_b) \sim \mathcal{D}$}
        \STATE Extract task-relevant features: $h_i = f_\theta(I_i, T_i, C_i)$
        \STATE Generate quality judgment: $y_i = \text{Head}(h_i)$
        \STATE Compute utility loss: $\mathcal{L}_{\text{utility}} = -\corr(y_i, P_i)$
        \STATE Compute explanation loss: $\mathcal{L}_{\text{expl}} = \simfunc(\text{Gen}(h_i), E_i)$
        \STATE Sample preference pairs $(I_1, I_2)$ where $P_1 > P_2$
        \STATE Compute DPO loss: $\mathcal{L}_{\text{DPO}}$ using \autoref{eq:dpo}
        \STATE Total loss: $\mathcal{L} = \lambda_1\mathcal{L}_{\text{utility}} + \lambda_2\mathcal{L}_{\text{expl}} + \lambda_3\mathcal{L}_{\text{DPO}}$
        \STATE Update: $\theta \leftarrow \theta - \alpha \nabla_\theta \mathcal{L}$
    \ENDFOR
\ENDFOR
  \end{algorithmic}
\end{algorithm}

The training procedure begins by initializing model parameters from a pre-trained MLLM, leveraging transfer learning from large-scale vision-language pretraining. During training, batches of images paired with task specifications and context are processed, extracting task-relevant features through multimodal encoding of $(I_i, T_i, C_i)$. This conditioning on task and context enables task-aware representations that differ fundamentally from traditional IQA methods that process images in isolation.

Feature extraction conditions on $(I_i, T_i, C_i)$ rather than just $I_i$, enabling the model to learn task-dependent quality representations. For crack detection, the model emphasizes edge clarity and surface visibility; for object manipulation, it prioritizes pose-relevant information. The utility loss uses correlation with actual task performance $P_i$ rather than human quality ratings, directly addressing evaluation misalignment by optimizing for task success. Explanation generation is integrated through similarity to expert annotations, ensuring quality judgments come with interpretable rationales. DPO operates on preference pairs determined by task performance rather than quality scores, aligning model preferences with actual utility. The model learns to prefer images that lead to task success, even if they score lower on traditional quality metrics.

\subsection{Evaluation Protocols}
\label{sec:appendix_eval}

We refer to \autoref{sec:eval} for the main evaluation principles. In the appendix, we note two practical additions: (i) include \emph{decision-centric} tests (e.g., whether the system correctly triggers retakes, repositioning, or abstention), and (ii) evaluate explanation fidelity via region-level grounding and expert verification \cite{lin2018koniq,kirstain2023pick}. Concretely, benchmark samples can include $(I,T,C)$ paired with task outcomes $P$ and expert rationales $E$, enabling evaluation of both utility prediction and rationale faithfulness.

\subsection{Deployment Strategies}

Deployment requires lightweight, real-time systems that integrate with decision-making pipelines. Knowledge distillation can compress large MLLMs into efficient models \cite{zhang2024embodied}. Specifically, a large teacher MLLM generates quality judgments and explanations, while a smaller student model learns to replicate these outputs with reduced computational cost. Quantization and hardware-aware neural architecture search can further optimize for edge devices, enabling real-time inference on resource-constrained platforms.

In UAV navigation \cite{abedin2020data}, AR inspection \cite{lv2025ar}, and robot perception \cite{nguyen2024mcd}, IQA agents must provide low-latency feedback, support user interaction, and operate within hardware constraints. UAV systems require energy-efficient navigation optimization \cite{abedin2020data}, while AR-assisted assembly methods \cite{lv2025ar} need real-time quality assessment for manufacturing tasks. Robot perception systems \cite{nguyen2024mcd} benefit from diverse multi-campus datasets that capture real-world variability. Integration with control systems enables closed-loop optimization, transforming IQA from post-hoc analysis to real-time perceptual support. For example, a drone's IQA agent can trigger image retaking when quality is insufficient for the current mission, or adjust flight parameters to improve image quality for specific tasks.

\section{Preliminary Evaluation}
\label{app:preliminary_evaluation}
\label{evaluation}


\subsection{Baselines}
In our experiments, we compare the proposed embodied IQA agent with the following state-of-the-art MLLM-based IQA baselines:
\begin{itemize}
    \item \textbf{Q-Insight}~\cite{li2025qinsightunderstandingimagequality} introduces a Reinforcement Learning (RL) based framework for image quality understanding. It utilizes Group Relative Policy Optimization (GRPO) to jointly optimize score regression and degradation perception tasks using carefully designed reward functions, demonstrating strong visual reasoning capabilities with limited data.

    \item \textbf{Q-SIT}~\cite{zhang2025teachinglmmsimagequality} proposes a unified framework capable of performing both image quality scoring and interpreting simultaneously. It incorporates a \emph{Scoring \& Interpreting Balance Strategy} that efficiently determines the optimal data mixture ratio via a lightweight proxy model to mitigate task interference and enhance cross-task knowledge transfer.

    \item \textbf{DeQA-Score}~\cite{you2025teachinglargelanguagemodels} focuses on accurate score regression by addressing the mismatch between continuous quality scores and discrete MLLM outputs. It employs a distribution-based soft label to preserve score distribution characteristics and introduces a fidelity loss based on Thurstone's model to capture intra-dataset relationships during multi-dataset co-training.
\end{itemize}

\subsection{Metrics}

To quantitatively evaluate image quality, we employ a diverse set of metrics ranging from traditional signal-fidelity measures to state-of-the-art deep learning-based perceptual models.

\subsubsection{Full-Reference (FR) IQA Metrics}
These metrics compare the distorted image $x$ against the pristine ground-truth image $y$.
\begin{itemize}
    \item \textbf{MSE (Mean Squared Error):} 
    MSE measures the average squared difference between the estimated values and the actual value. It represents the signal fidelity in terms of pixel-wise error.
    \begin{equation}
        \text{MSE} = \frac{1}{N} \sum_{i=1}^{N} (x_i - y_i)^2
    \end{equation}
    where $N$ is the total number of pixels. A lower MSE indicates less error and higher similarity to the reference.
    \item \textbf{PSNR (Peak Signal-to-Noise Ratio) :} 
    PSNR is derived from MSE and expresses the ratio between the maximum possible power of a signal and the power of corrupting noise. It is widely used to quantify reconstruction quality.
    \begin{equation}
        \text{PSNR} = 10 \cdot \log_{10} \left( \frac{L^2}{\text{MSE}} \right)
    \end{equation}
    where $L$ is the maximum possible pixel value (e.g., 255 for 8-bit images). Higher PSNR values imply better quality.
    \item \textbf{SSIM (Structural Similarity Index Measure) ~\cite{wang2004image}:} 
    SSIM mimics the human visual system (HVS) by evaluating structural information degradation rather than pixel-wise error. It assesses three components: luminance ($l$), contrast ($c$), and structure ($s$).
    \begin{equation}
        \text{SSIM}(x, y) = [l(x,y)]^\alpha \cdot [c(x,y)]^\beta \cdot [s(x,y)]^\gamma
    \end{equation}
    Values range from 0 to 1, where 1 indicates perfect structural identity.
    \item \textbf{LPIPS (Learned Perceptual Image Patch Similarity) ~\cite{zhang2018unreasonable}:} 
    LPIPS evaluates the distance between image patches in the deep feature space of a pre-trained convolutional neural network (e.g., VGG or AlexNet). It correlates better with human perceptual judgment than traditional metrics like MSE/PSNR.
    \begin{equation}
        d(x, y) = \sum_l \frac{1}{H_l W_l} \sum_{h,w} || w_l \odot (\hat{y}^l_{hw} - \hat{y}^l_{0hw}) ||_2^2
    \end{equation}
    Lower LPIPS scores indicate that the distorted image is perceptually closer to the reference.
    \item \textbf{DISTS (Deep Image Structure and Texture Similarity) ~\cite{ding2020image}:} 
    DISTS is a deep learning-based metric that specifically focuses on texture and structure invariance. It combines structural similarity measurements with deep feature maps to be robust against texture substitution, aligning closely with human texture perception. Lower values indicate higher similarity.
\end{itemize}

\subsubsection{No-Reference (NR) IQA Metrics}
These metrics assess image quality without access to a ground-truth reference, relying on learned statistical models of natural image statistics.
\begin{itemize}
    \item \textbf{NIQE (Natural Image Quality Evaluator) ~\cite{mittal2012making}:} 
    NIQE is a completely blind IQA model based on constructing a ``quality-aware'' collection of statistical features derived from a corpus of natural, undistorted images. It measures the distance between the multivariate Gaussian (MVG) model of the test image and the MVG model of natural images. Lower scores indicate better perceptual quality, i.e., closer alignment with natural image statistics.
    \item \textbf{BRISQUE (Blind/Referenceless Image Spatial Quality Evaluator) ~\cite{mittal2012no}:} 
    BRISQUE operates in the spatial domain using scene statistics of locally normalized luminance coefficients. It quantifies the loss of naturalness using a support vector regression (SVR) model trained on human-rated datasets. Lower scores indicate better quality.
    \item \textbf{MANIQA (Multi-dimension Attention Network for IQA) ~\cite{yang2022maniqa}:} 
    MANIQA is a state-of-the-art deep learning method that utilizes a Multi-dimension Attention Network. It leverages Vision Transformer (ViT) features and applies channel-wise and spatial attention mechanisms to capture interaction across different regions, predicting a quality score that aligns with human mean opinion scores (MOS). Higher scores indicate better quality.
    \item \textbf{HyperIQA (Hyper-network for IQA) ~\cite{su2020blindly}:} 
    HyperIQA addresses the content-dependency problem in blind IQA. It uses a self-adaptive hyper-network to separate content understanding from quality prediction, generating specific weight parameters for different image contents to predict the final quality score. Higher scores indicate better quality.
\end{itemize}

\subsection{Task Scenarios}
\noindent\textbf{Task 1 (Robotic Grasping):} \textit{In a precise manipulation scenario, the objective is to resolve the 6D pose of the target object and target coordinates, ensuring visual cues are robust against quantization artifacts for successful grasp execution.}

\vspace{0.5em}

\begin{figure}[htbp]
    \centering
    \begin{subfigure}[t]{0.35\textwidth}
        \centering
        \includegraphics[width=\linewidth, height=0.8\linewidth, keepaspectratio]{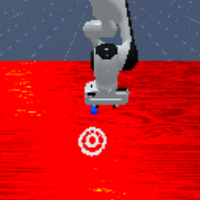}

        \caption{Input Image (Distorted)}
        \label{fig:task1:distort}
    \end{subfigure}
    \hspace{0.5cm} 
    \begin{subfigure}[t]{0.35\textwidth}
        \centering
        \includegraphics[width=\linewidth, height=0.8\linewidth, keepaspectratio]{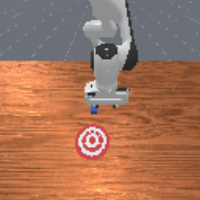}
        \caption{Reference Image}
        \label{fig:task1:ref}
    \end{subfigure}
    
    \caption{Visual visualization for Task 1 (Robotic Grasping). Left: The distorted input; Right: The reference ground truth.}
    \label{fig:app:task1_visual}
\end{figure}

\vspace{0.25 em}
\noindent\textbf{Detailed Results for Task 1:}
\vspace{2 em}
\begin{tcolorbox}[
    breakable, 
    title=\textbf{Traditional \& MLLM-based Methods},
    colback=white, 
    colframe=gray!80!black,
    coltitle=white,
    arc=0 pt, 
    outer arc=0 pt,
    before skip=0 pt,
    after skip=0 pt
]
\begin{Verbatim}[breaklines=true, breakanywhere=true, fontsize=\footnotesize, commandchars=\\\{\}, breaksymbolleft={}, mathescape=true]
\textbf{[Traditional IQA Scores]}
 % Full-Reference (FR-IQA)
 MSE: 5925.8981 ($\downarrow$ Low)
 PSNR: 10.4033 ($\uparrow$ High)
 SSIM [0, 1]: 0.1505 ($\uparrow$ High)
 LPIPS [0, 1]: 0.7858 ($\downarrow$ Low)
 DISTS [0, 1]: 0.4567 ($\downarrow$ Low)

 % No-Reference (NR-IQA)
 NIQE: 8.2755 ($\downarrow$ Low)
 BRISQUE: 22.9999 ($\downarrow$ Low)
 MANIQA [0, 1]: 0.1501 ($\uparrow$ High)
 HyperIQA [0, 1]: 0.2694 ($\uparrow$ High)

\textbf{[MLLM-based IQA methods]}
 Q-Insight:
  Model Response:
  <think> 
  This image appears to be a digitally generated or pixelated image, possibly a sprite from a video game or an illustration, given the style. The image contains recognizable elements such as a humanoid figure and a target, suggesting it's a constructed or intended image. There's a level of detail present, but the pixelation and resolution give the image a somewhat retro or stylized aesthetic. The colors are distinct, and there's a clear contrast between elements, which aids in visibility and comprehensibility. Overall, it's clear and legible, but lacks fine detail. Judging by the pixelation, lack of fine detail, and resolution, the image quality isn't quite at a high level, but the content is still understandable and appealing, especially for what it seems to be.
  </think>
  <answer>
  {"rating": 3.00}
  </answer>
  
 Q-SIT: 4.542420948398246
 
 DeQA-Score: 2.0176
\end{Verbatim}
\end{tcolorbox}

\vspace{-4 em}
\begin{tcolorbox}[
    breakable, 
    title=\textbf{Embodied IQA Agent},
    colback=teal!5!white,     
    colframe=teal!80!black,   
    coltitle=white,
    fonttitle=\bfseries,     
    arc=0 pt, 
    outer arc=0 pt,
    before skip=0 pt,
    after skip=0 pt
]
\begin{Verbatim}[breaklines=true, breakanywhere=true, fontsize=\footnotesize, commandchars=\\\{\}, breaksymbolleft={}, mathescape=true]
\textbf{ 1. Context Adaptation Report}
 The aerial perspective provides a wide field of view with neutral-toned background elements (gray asphalt, green foliage). The lighting is diffuse (overcast), which minimizes harsh specular reflections that might otherwise distort color perception. While there is a slight global softness typical of high-altitude surveillance footage, the color saturation of red remains a dominant visual feature in the scene.

\textbf{ 2. Task-Grounded Explanation}
 Localization: The target red vehicle is located in the center-right quadrant, specifically on the paved shoulder/sidewalk area adjacent to the row of trees.
 Impact Analysis: The visual conditions are optimal for the specific detection task. The red color signature exhibits high chromatic contrast against both the gray road surface and the green grass, allowing for immediate color-based thresholding. Furthermore, despite the high altitude, the vehicle's geometric boundaries are sharp enough to allow a detection algorithm to generate a precise bounding box without confusion from neighboring objects or background textures.

\textbf{ 3. Embodied Integration (Control Decision Support)}
 Perception Action: The vision system should prioritize a color-space segmentation (e.g., HSV filtering) to isolate the target. The current signal-to-noise ratio is high enough to support automated tracking with minimal risk of identity switching or loss of lock.
 Confidence Threshold: High. No additional image enhancement or altitude adjustment is required for successful task execution.

\textbf{ 4. Embodied Utility Score: 5.0}
\end{Verbatim}
\end{tcolorbox}

\vspace{-6 em}
\captionof{lstlisting}{Quantitative and reasoning results for Task 1.}
\label{lst:app:task1}

\vspace{1 em}

\noindent\textbf{Task 2 (Aerial Detection):} \textit{In a high-altitude surveillance scenario, the objective is to locate and bound a specific target red vehicle based on its chromatic signature amidst neutral background elements.}

\vspace{0.5 em}

\begin{center}
    \includegraphics[width=0.6\textwidth]{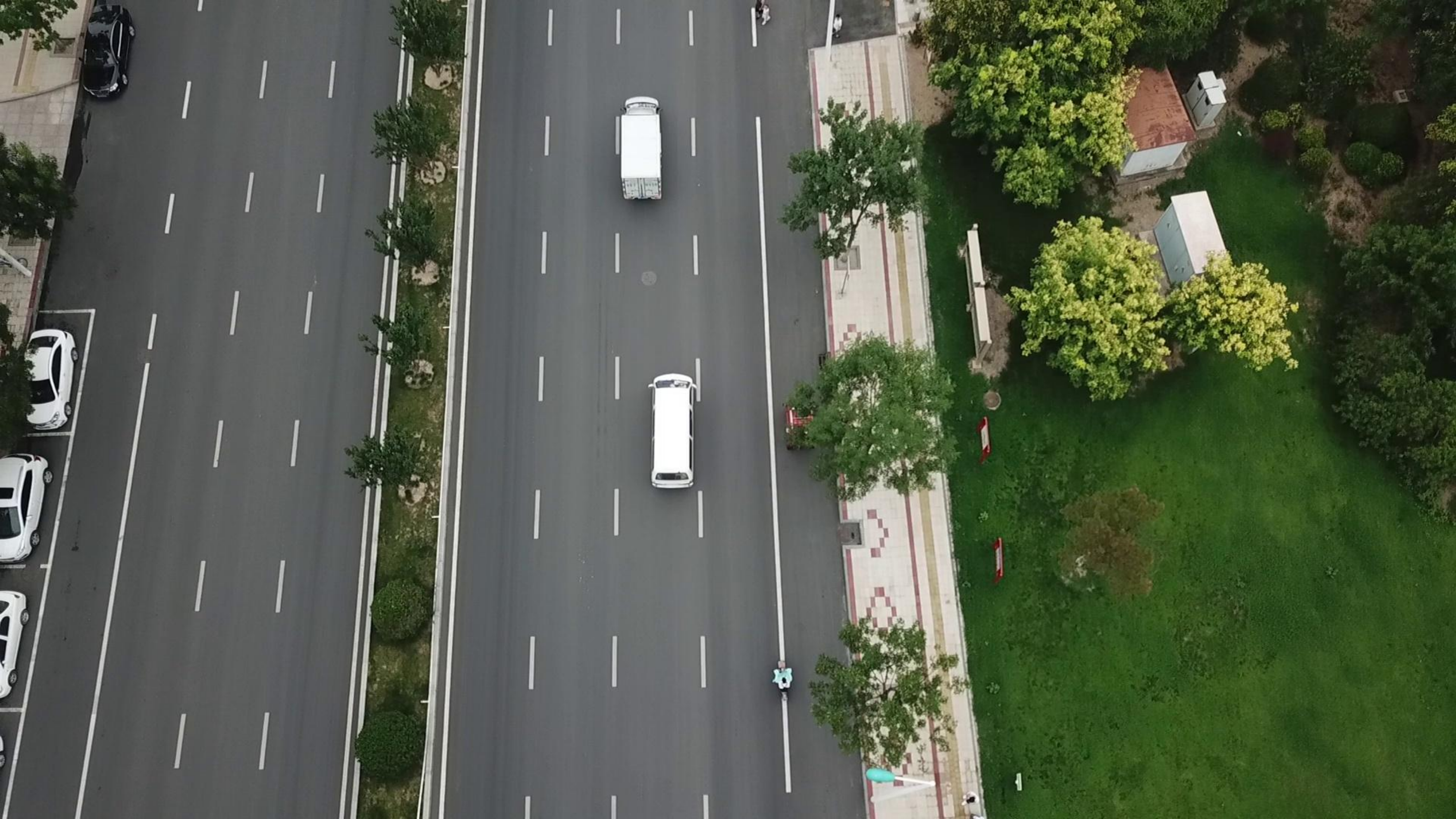}
    \captionof{figure}{Input Image for Task 2.}
    \label{fig:app:task2}
\end{center}

\vspace{0.25em}

\noindent\textbf{Detailed Results for Task 2:}

\vspace{1.5 em}

\begin{tcolorbox}[
    breakable, 
    title=\textbf{Traditional \& MLLM-based Methods},
    colback=white, 
    colframe=gray!80!black,
    coltitle=white,
    arc=0pt, 
    outer arc=0pt,
    before skip=0 pt,
    after skip=0 pt
]
\begin{Verbatim}[breaklines=true, breakanywhere=true, fontsize=\footnotesize, commandchars=\\\{\}, breaksymbolleft={}, mathescape=true]
\textbf{[Traditional IQA Scores]}
 NIQE: 3.0114 ($\downarrow$ Low)
 BRISQUE: 1.8237 ($\downarrow$ Low)
 MANIQA [0,1]: 0.2950 ($\uparrow$ High)
 HyperIQA [0,1]: 0.5463 ($\uparrow$ High)

\textbf{[MLLM-based IQA methods]}
 Q-Insight:
  Model Response:
  <think> 
  This is a picture that has a good level of detail and clear lines, as well as good contrast between the roads, vehicles, and surrounding greenery. The colors are balanced, and the subject matter is in focus. The composition provides a clear view of the subjects, though there is a slight graininess or noise in the image, which slightly decreases the sharpness but does not significantly impact the overall image quality. There is a lack of motion blur, which is evident in the car and vehicles, giving it a clear and legible appearance.
  </think>
  <answer>
  {"rating": 3.80}
  </answer>
  
 Q-SIT: 4.542420948398246
 
 DeQA-Score: 2.0176
\end{Verbatim}
\end{tcolorbox}

\vspace{2 em}
\begin{tcolorbox}[
    breakable, 
    title=\textbf{Embodied IQA Agent},
    colback=teal!5!white,     
    colframe=teal!80!black,   
    coltitle=white,
    fonttitle=\bfseries,     
    arc=0pt, 
    outer arc=0pt,
    before skip=0 pt,
    after skip=0 pt
]
\begin{Verbatim}[breaklines=true, breakanywhere=true, fontsize=\footnotesize, commandchars=\\\{\}, breaksymbolleft={}, mathescape=true]
\textbf{ 1. Context Adaptation Report}
 The aerial perspective provides a wide field of view with neutral-toned background elements (gray asphalt, green foliage). The lighting is diffuse (overcast), which minimizes harsh specular reflections that might otherwise distort color perception. While there is a slight global softness typical of high-altitude surveillance footage, the color saturation of red remains a dominant visual feature in the scene.

\textbf{ 2. Task-Grounded Explanation}
 Localization: The target red vehicle is located in the center-right quadrant, specifically on the paved shoulder/sidewalk area adjacent to the row of trees.
 Impact Analysis: The visual conditions are optimal for the specific detection task. The red color signature exhibits high chromatic contrast against both the gray road surface and the green grass, allowing for immediate color-based thresholding. Furthermore, despite the high altitude, the vehicle's geometric boundaries are sharp enough to allow a detection algorithm to generate a precise bounding box without confusion from neighboring objects or background textures.

\textbf{ 3. Embodied Integration (Control Decision Support)}
 Perception Action: The vision system should prioritize a color-space segmentation (e.g., HSV filtering) to isolate the target. The current signal-to-noise ratio is high enough to support automated tracking with minimal risk of identity switching or loss of lock.
 Confidence Threshold: High. No additional image enhancement or altitude adjustment is required for successful task execution.

\textbf{ 4. Embodied Utility Score: 5.0}
\end{Verbatim}
\end{tcolorbox}

\vspace{0.8 em}

\captionof{lstlisting}{Quantitative and reasoning results for Task 2.}
\label{lst:app:task2}

\vspace{1 em}

\noindent\textbf{Task 3 (Auto. Driving):} \textit{In a dynamic roadway environment, the objective is to evaluate the visual utility for safety-critical perception (e.g., lane topology, signal states) and identify fatal navigational violations such as wrong-way driving.}

\vspace{0.5 em}

\begin{center}
    \includegraphics[width=0.5\textwidth]{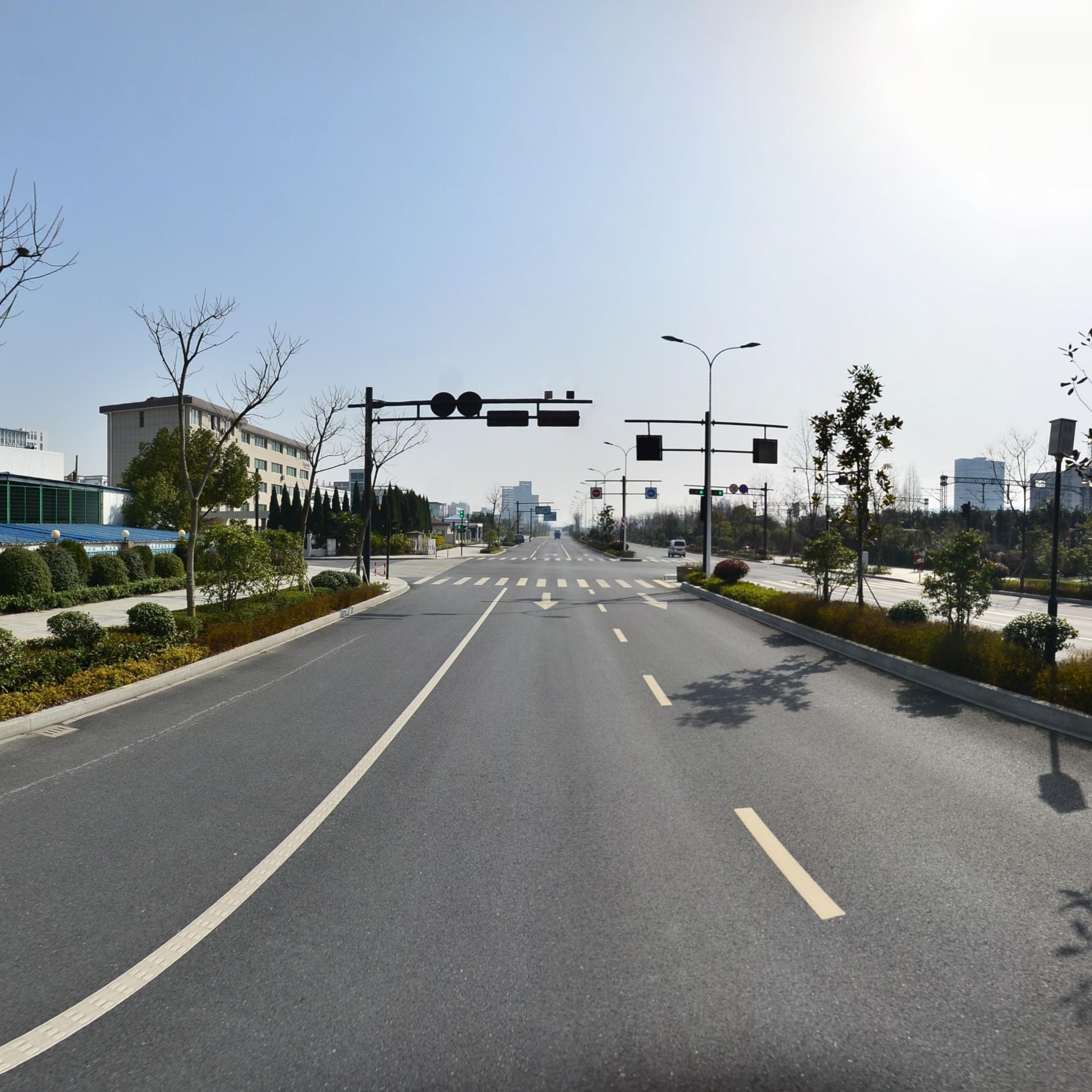}
    \captionof{figure}{Input Image for Task 3.}
    \label{fig:app:task3}
\end{center}
\vspace{0.25 em}

\noindent\textbf{Detailed Results for Task 3:}
\vspace{2 em}
\begin{tcolorbox}[
    breakable, 
    title=\textbf{Traditional \& MLLM-based Methods},
    colback=white, 
    colframe=gray!80!black,
    coltitle=white,
    arc=0pt, 
    outer arc=0pt,
    before skip=0 pt,
    after skip=0 pt
]
\begin{Verbatim}[breaklines=true, breakanywhere=true, fontsize=\footnotesize, commandchars=\\\{\}, breaksymbolleft={}, mathescape=true]
\textbf{[Traditional IQA Scores]}
 NIQE: 3.1552 ($\downarrow$ Low)
 BRISQUE: 16.7502 ($\downarrow$ Low)
 MANIQA [0,1]: 0.3850 ($\uparrow$ High)
 HyperIQA [0,1]: 0.5344 ($\uparrow$ High)

\textbf{[MLLM-based IQA methods]}
 Q-Insight:
   Model Response:
   <think> 
   This picture is clear and detailed, providing a sense of depth and perspective, as well as good lighting and composition. The image is well-balanced, with a clear focus on the intersection, and it is free from blurriness or distracting elements. The color balance appears natural and vibrant. The photograph effectively captures the environment, suggesting good image quality that is suitable for its subject.
   </think>
   <answer>
   {"rating": 4.17}
   </answer>
  
 Q-SIT: 4.8053
 
 DeQA-Score: 4.4023
\end{Verbatim}
\end{tcolorbox}

\vspace{2 em}

\begin{tcolorbox}[
    breakable, 
    title=\textbf{Embodied IQA Agent},
    colback=teal!5!white,     
    colframe=teal!80!black,   
    coltitle=white,
    fonttitle=\bfseries,     
    arc=0pt, 
    outer arc=0pt,
    before skip=0 pt,
    after skip=0 pt
]
\begin{Verbatim}[breaklines=true, breakanywhere=true, fontsize=\footnotesize, commandchars=\\\{\}, breaksymbolleft={}, mathescape=true]
\textbf{1. Context Adaptation Report}
The environment presents high-contrast lighting with a direct solar source in the upper right. Sharp shadows from roadside trees are cast across the asphalt. Despite these high-frequency visual interruptions, sensor reliability is excellent; the dynamic range is sufficient to maintain high edge-contrast for lane markings within the shadows. No lens flare or atmospheric occlusion (fog/rain) impairs the distant gantry or ground instructions.

\textbf{2. Task-Grounded Explanation}
Visual Robustness: Image quality is extremely high. The resolution and contrast support high-confidence detection of fine-grained semantics, including ground arrow contours and the mechanical structure of distant signal housings.
Compliance Check:
Directional Arrows: A clear white arrow on the road surface at the intersection entrance is oriented facing the vehicle.
Signal Orientation: The primary overhead traffic signals are seen from the rear (housing side), indicating they are intended for traffic traveling in the opposite direction.
Lane Logic: The vehicle is positioned to the left of a yellow dashed line; in standard right-hand traffic jurisdictions, this confirms the vehicle has crossed into the opposing flow of traffic.
Finding: CRITICAL STATE DETECTED - Wrong-Way Driving.
Safety Conclusion: While the visual data is "perfect" in terms of clarity, it provides undeniable evidence of a fatal navigational violation. The situation is inherently unsafe for continued autonomous operation.

\textbf{3. Embodied Integration (Control Decision Support)}
Driving Policy: Immediate Handover (Critical Failure/Violation).
Reasoning: The perception system has identified a high-confidence "Wrong-Way" scenario. Based on the Safety Veto logic, the viability of continuous autonomy is zero. The system must immediately trigger an emergency alert and hand over control to a human operator or execute a safe-stop maneuver to prevent a head-on collision.

\textbf{4. Embodied Utility Score}
1.0
(Justification: Even though the image is crystal clear (Sensor Quality = 5.0), the detection of a Critical Violation (Wrong-Way) mandates a score of 1.0. The data confirms that the current trajectory is illegal and life-threatening.)
\end{Verbatim}
\end{tcolorbox}

\vspace{0.8 em}

\captionof{lstlisting}{Quantitative and reasoning results for Task 3.}
\label{lst:app:task3}

\vspace{1 em}

\noindent\textbf{Task 4 (Authenticity Check):} \textit{In a mixed-reality power grid inspection scenario, the objective is to filter out invalid synthetic data by detecting physical inconsistencies (e.g., floating artifacts without gravity support) that violate environmental realism.}

\vspace{0.5em}

\begin{center}
    \includegraphics[width=0.6\textwidth]{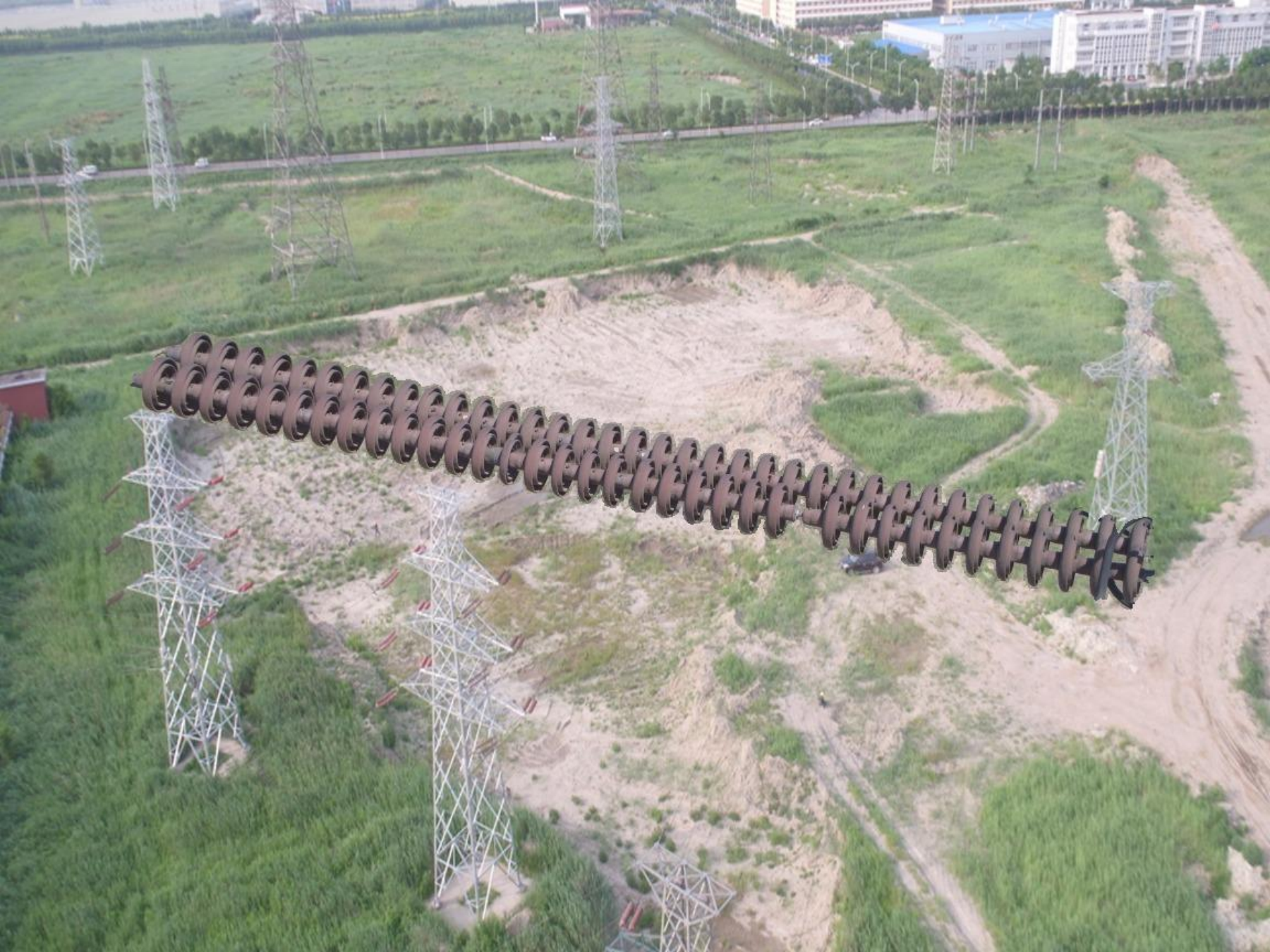}
    \captionof{figure}{Input Image for Task 4.}
    \label{fig:app:task4}
\end{center}

\vspace{0.05em}
\noindent\textbf{Detailed Results for Task 4:}
\vspace{2 em}
\begin{tcolorbox}[
    breakable, 
    title=\textbf{Traditional \& MLLM-based Methods},
    colback=white, 
    colframe=gray!80!black,
    coltitle=white,
    arc=0pt, 
    outer arc=0pt,
    before skip=0 pt,
    after skip=0 pt
]
\begin{Verbatim}[breaklines=true, breakanywhere=true, fontsize=\footnotesize, commandchars=\\\{\}, breaksymbolleft={}, mathescape=true]
\textbf{[Traditional IQA Scores]}
 NIQE: 3.1446 ($\downarrow$ Low)
 BRISQUE: 24.9523 ($\downarrow$ Low)
 MANIQA [0,1]: 0.3698 ($\uparrow$ High)
 HyperIQA [0,1]: 0.4684 ($\uparrow$ High)

\textbf{[MLLM-based IQA methods]}
 Q-Insight:
   Model Response:
   <think> 
   This image appears to be an outdoor scene captured with a good level of detail and focus. The resolution seems adequate, with distinguishable details on the subject and clear outlines of the landscape. The image is not particularly vibrant in color, which could indicate a more natural or neutral environment. The photo shows some slight grain or noise, which is common in some types of image capture, but it does not significantly impact the overall composition or readability.
   </think>
   <answer>
   {"rating": 3.56}
   </answer>
  
 Q-SIT: 4.0642
 
 DeQA-Score: 3.6074
\end{Verbatim}
\end{tcolorbox}

\vspace{2 em}

\begin{tcolorbox}[
    breakable, 
    title=\textbf{Embodied IQA Agent},
    colback=teal!5!white,     
    colframe=teal!80!black,   
    coltitle=white,
    fonttitle=\bfseries,     
    arc=0pt, 
    outer arc=0pt,
    before skip=0 pt,
    after skip=0 pt
]
\begin{Verbatim}[breaklines=true, breakanywhere=true, fontsize=\footnotesize, commandchars=\\\{\}, breaksymbolleft={}, mathescape=true]
\textbf{1. Context Adaptation Report}
The image is a clear instance of a synthetic simulation or a "phantom" composite. While the background (fields, roads, and distant towers) appears to be a real-world aerial capture, the primary subject-the insulator string-exhibits massive physics violations. There is a total absence of mechanical support (it is floating in mid-air), no interaction with environmental lighting (it casts no shadow on the ground or nearby towers), and it possesses "jagged" edge artifacts that indicate a poor digital insertion.
Verdict: Synthetic / Fake Data.

\textbf{2. Task-Grounded Explanation}
Authenticity Verification: The insulator string is floating without any physical connection to the tower arms or transmission cables. It violates the law of gravity and the fundamental structural logic of power grid infrastructure.
Defect Analysis: Because the object is a digital "phantom" and not a real physical component in the scene, performing a health monitoring assessment is scientifically invalid. Any apparent "defects" would be artifacts of the rendering/compositing process rather than actual grid hazards.
Inspection Conclusion: Invalid Data. This frame is a "hallucination" or a test case designed to verify the perception system's ability to filter out non-existent objects.

\textbf{3. Embodied Integration (Control Decision Support)}
Recommended Action: Discard/Ignore.
Reasoning: The data is non-authentic. Recording this as a "healthy" or "defective" component would corrupt the grid's asset database with false information. The system must flag this frame as a sensor/processing anomaly and proceed to the next valid frame.

\textbf{4. Embodied Utility Score}
1.0
(Justification: The image is a synthetic "phantom" that provides zero utility for actual grid health monitoring. It is correctly identified as invalid data.)
\end{Verbatim}
\end{tcolorbox}

\vspace{0.8 em}

\captionof{lstlisting}{Quantitative and reasoning results for Task 4.}
\label{lst:app:task4}

\vspace{1 em}

\noindent\textbf{Task 5 (Defect Detection):} \textit{In a verified real-world transmission line inspection, the objective is to assess the structural health of insulators and identify specific physical hazards, such as foreign objects (e.g., bird nests) or component breakage.}

\vspace{0.5em}

\begin{center}
    \includegraphics[width=0.5\textwidth]{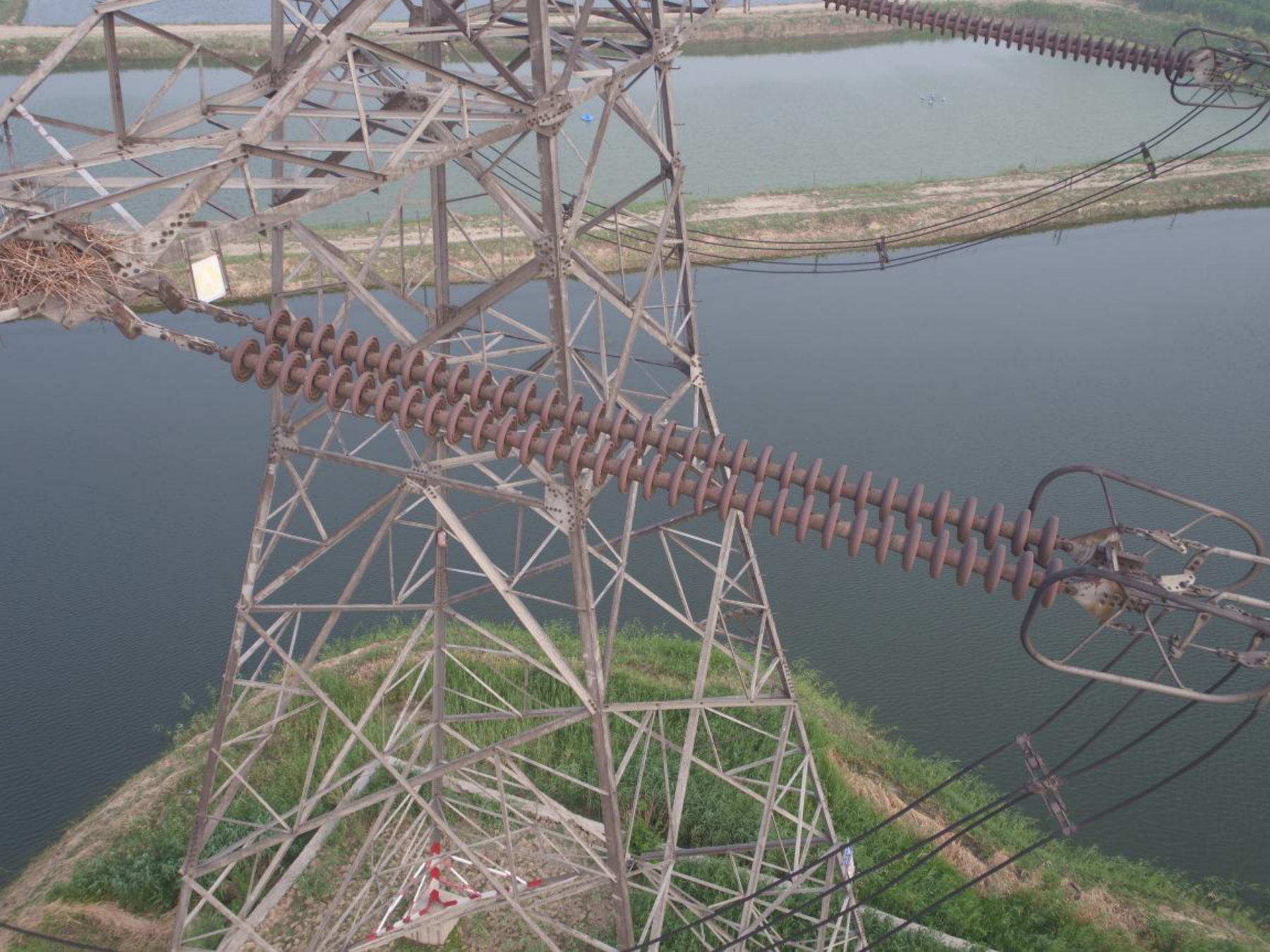}
    \captionof{figure}{Input Image for Task 5.}
    \label{fig:app:task5}
\end{center}
\vspace{0.25em}

\noindent\textbf{Detailed Results for Task 5:}
\vspace{2 em}
\begin{tcolorbox}[
    breakable, 
    title=\textbf{Traditional \& MLLM-based Methods},
    colback=white, 
    colframe=gray!80!black,
    coltitle=white,
    arc=0pt, 
    outer arc=0pt,
    before skip=0 pt,
    after skip=0 pt
]
\begin{Verbatim}[breaklines=true, breakanywhere=true, fontsize=\footnotesize, commandchars=\\\{\}, breaksymbolleft={}, mathescape=true]
\textbf{[Traditional IQA Scores]}
 NIQE: 2.6471 ($\downarrow$ Low)
 BRISQUE: 14.0626 ($\downarrow$ Low)
 MANIQA [0,1]: 0.5057 ($\uparrow$ High)
 HyperIQA [0,1]: 0.6792 ($\uparrow$ High)

\textbf{[MLLM-based IQA methods]}
 Q-Insight:
   Model Response:
   <think> 
   This picture appears to be clear with good focus, and the details of the power tower and the surrounding environment are well-preserved. There's a slight lack of sharpness, but the lighting is even, and the colors are natural. No obvious issues in color balance, exposure, or distortion. Overall, the picture seems balanced and the subject is well-defined within the frame. 
   </think>
   <answer>
   {"rating": 3.87}
   </answer>
  
 Q-SIT: 4.4320
 
 DeQA-Score: 4.0547
\end{Verbatim}
\end{tcolorbox}

\vspace{2 em}
\begin{tcolorbox}[
    breakable, 
    title=\textbf{Embodied IQA Agent},
    colback=teal!5!white,     
    colframe=teal!80!black,   
    coltitle=white,
    fonttitle=\bfseries,     
    arc=0pt, 
    outer arc=0pt,
    before skip=0 pt,
    after skip=0 pt
]
\begin{Verbatim}[breaklines=true, breakanywhere=true, fontsize=\footnotesize, commandchars=\\\{\}, breaksymbolleft={}, mathescape=true]
\textbf{1. Context Adaptation Report}
The scene exhibits high environmental realism. The lighting is diffuse and consistent with an overcast sky, casting soft, natural shadows across the lattice structure. Material textures-specifically the oxidation/weathering on the galvanized steel and the organic debris of the bird nest-show high-fidelity details that match real-world physical aging. The background (water ponds and levees) displays realistic atmospheric perspective and water-surface ripples. No "floating" artifacts or compositing halos are present.
Verdict: Real-world capture.

\textbf{2. Task-Grounded Explanation}
Authenticity Verification: The image is physically valid. The insulator strings are correctly under tension and mechanically anchored to the tower's cross-arms using standard hardware (shackles and ball-eye links). The spatial relationship between the drone's perspective, the tower, and the ground is geometrically consistent.
Defect Analysis: A significant Foreign Object Hazard (Bird Nest) is identified on the left-hand cross-arm, located directly at the critical junction where the insulator string attaches to the tower. While the insulator discs themselves appear structurally intact (no visible broken sheds), the proximity of the nest to high-voltage components poses a high risk of flashover or fire during damp conditions.
Inspection Conclusion: Defect Hazard. This is a valid, high-quality inspection frame that has successfully captured a maintenance-critical anomaly.

\textbf{3. Embodied Integration (Control Decision Support)}
Recommended Action: Trigger Defect Alert.
Reasoning: The detection of a bird nest in a high-voltage clearance zone is a primary maintenance trigger. The visual data is clear and unambiguous, providing sufficient evidence for a maintenance crew to be dispatched for removal.

\textbf{4. Embodied Utility Score}
3.0
(Justification: This score represents a real, clear image where a specific defect/foreign object has been detected, necessitating system intervention.)
\end{Verbatim}
\end{tcolorbox}

\vspace{0.8 em}

\captionof{lstlisting}{Quantitative and reasoning results for Task 5.}
\label{lst:app:task5}

\vspace{1 em}

\noindent\textbf{Task 6 (Indoor Navigation):} \textit{In a resource-constrained indoor environment (e.g., dimly lit corridors), the objective is to determine navigational actionability by verifying the visibility of essential geometric boundaries and obstacles despite severe sensor noise.}

\vspace{0.5em}

\begin{center}
    \includegraphics[width=0.5\textwidth]{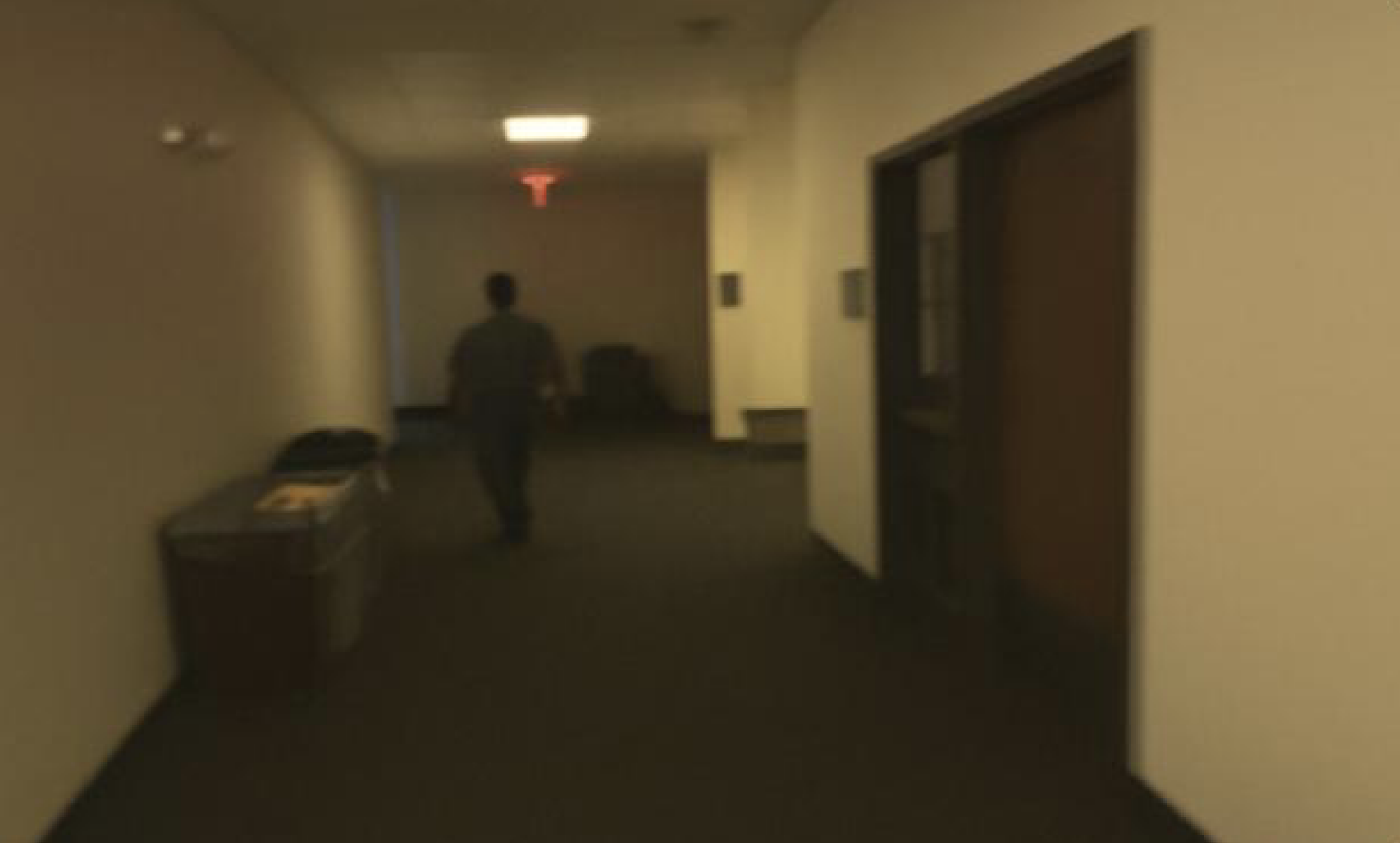}
    \captionof{figure}{Input Image for Task 6.}
    \label{fig:app:task6}
\end{center}
\vspace{0.25em}

\noindent\textbf{Detailed Results for Task 6:}
\vspace{2 em}
\begin{tcolorbox}[
    breakable, 
    title=\textbf{Traditional \& MLLM-based Methods},
    colback=white, 
    colframe=gray!80!black,
    coltitle=white,
    arc=0pt, 
    outer arc=0pt,
    before skip=0 pt,
    after skip=0 pt
]
\begin{Verbatim}[breaklines=true, breakanywhere=true, fontsize=\footnotesize, commandchars=\\\{\}, breaksymbolleft={}, mathescape=true]
\textbf{[Traditional IQA Scores]}
 NIQE: 11.8701 ($\downarrow$ Low)
 BRISQUE: 79.0786 ($\downarrow$ Low)
 MANIQA [0,1]: 0.2994 ($\uparrow$ High)
 HyperIQA [0,1]: 0.3356 ($\uparrow$ High)

\textbf{[MLLM-based IQA methods]}
 Q-Insight:
   Model Response:
   <think> 
   This picture appears to be of a hallway or corridor in an office or a similar building. The resolution is not very high, and there's a slight lack of sharpness, which may indicate motion blur or poor image focus. Lighting seems adequate but does not brighten the image enough, potentially making the subject slightly darker or obscured. It's slightly blurry, potentially because it was taken with a handheld device, or there was motion when the picture was taken, giving it a grainy texture that could mask further detail. The person in the picture is not fully defined, and there isn't a high level of fine detail or fine-textured elements in the scene that would be expected in a high-quality image.
   </think>
   <answer>
   {"rating": 2.30}
   </answer>
  
 Q-SIT: 0.5875
 
 DeQA-Score: 1.8262
\end{Verbatim}
\end{tcolorbox}

\vspace{2 em}
\begin{tcolorbox}[
    breakable, 
    title=\textbf{Embodied IQA Agent},
    colback=teal!5!white,     
    colframe=teal!80!black,   
    coltitle=white,
    fonttitle=\bfseries,     
    arc=0pt, 
    outer arc=0pt,
    before skip=0 pt,
    after skip=0 pt
]
\begin{Verbatim}[breaklines=true, breakanywhere=true, fontsize=\footnotesize, commandchars=\\\{\}, breaksymbolleft={}, mathescape=true]
\textbf{1. Context Adaptation Report}
The image is characterized by low light, significant sensor noise (grain), and noticeable motion blur. The dynamic range is poor, with the overhead light source causing localized blooming. However, despite being "low-fidelity" by aesthetic standards, the environmental geometry is not destroyed. The contrast between the light-colored walls and the dark-toned carpet remains sufficient for a standard Canny or Sobel edge-detection filter to identify the primary structural planes.

\textbf{2. Task-Grounded Explanation}
Geometry Check: The floor-wall boundaries are continuous and visible. The perspective lines of the corridor converge toward a clear vanishing point, which allows the navigation stack to maintain a reliable heading. Although the edges are "soft" due to blur, the transition from vertical to horizontal planes is semantically distinct.
Obstacle Detection:
Dynamic: The silhouette of the person walking ahead is high-contrast against the lighter background at the end of the hall, ensuring they are mapped as a dynamic obstacle in the occupancy grid.
Static: The trash bin on the left and the doorframe on the right have clear enough silhouettes for the robot to calculate a clearance buffer.
Navigability: There is a clear "traversable path" to the right of the bin and behind the pedestrian. The robot has enough visual data to calculate a local trajectory vector.

\textbf{3. Embodied Integration (Control Decision Support)}
Recommended Control State: Reduce Speed (Low Confidence).
Reasoning: The image is "Usable," but the low resolution and motion blur increase the margin of error for depth estimation. Reducing speed will allow the robot to perform temporal filtering (averaging multiple frames) to compensate for sensor noise and ensure that obstacle positions are stable before committing to a high-speed path.

\textbf{4. Embodied Utility Score}
3.5
(Justification: The data is noisy and dim, but the safety-critical information-the location of the floor, the walls, and the obstacles-is fully actionable. The robot can navigate safely, albeit with higher caution than in a well-lit environment.)
\end{Verbatim}
\end{tcolorbox}

\vspace{0.8 em}

\captionof{lstlisting}{Quantitative and reasoning results for Task 6.}
\label{lst:app:task6}

\vspace{1 em}

\subsection{Evaluation Results}

\input{tables/results}

\subsection{Additional Empirical Evidence on VisDrone}
\label{app:visdrone_empirical_evidence}

To complement the main evidence study, we provide additional VisDrone results that compare generic IQA baselines with the task-conditioned utility score, Ours-TCUS. The purpose of this appendix is not to claim a fully saturated benchmark, but to provide supporting evidence for the central hypothesis of this work: in embodied settings, image quality should be evaluated with respect to task utility rather than task-agnostic perceptual fidelity.

\paragraph{External baseline comparison.}
Table~\ref{tab:app_visdrone_external_baseline} reports the ability of different IQA scores to predict downstream task success on VisDrone. Generic MLLM-based IQA baselines, including Q-SiT, Q-Insight, and DeQA-Score-Mix3, remain weak predictors of task success, while Ours-TCUS achieves substantially stronger alignment across AUROC, average precision, and Spearman correlation.

\begin{table}[t]
\centering
\scriptsize
\setlength{\tabcolsep}{3.5pt}
\caption{Additional VisDrone evidence for predicting downstream task success. Higher is better for AUROC, AP, and Spearman correlation.}
\label{tab:app_visdrone_external_baseline}
\begin{tabular}{lccc}
\toprule
Method & AUROC $\uparrow$ & AP $\uparrow$ & $\rho$ $\uparrow$ \\
\midrule
Q-SiT & 0.510 & 0.632 & 0.017 \\
Q-Insight & 0.502 & 0.629 & 0.003 \\
DeQA-Score-Mix3 & 0.500 & 0.649 & 0.001 \\
Cognition-Consistency & 0.623 & 0.694 & 0.352 \\
\midrule
\textbf{Ours-TCUS} & \textbf{0.925} & \textbf{0.948} & \textbf{0.710} \\
\bottomrule
\end{tabular}
\end{table}

\begin{figure}[t]
\centering
\includegraphics[width=\columnwidth]{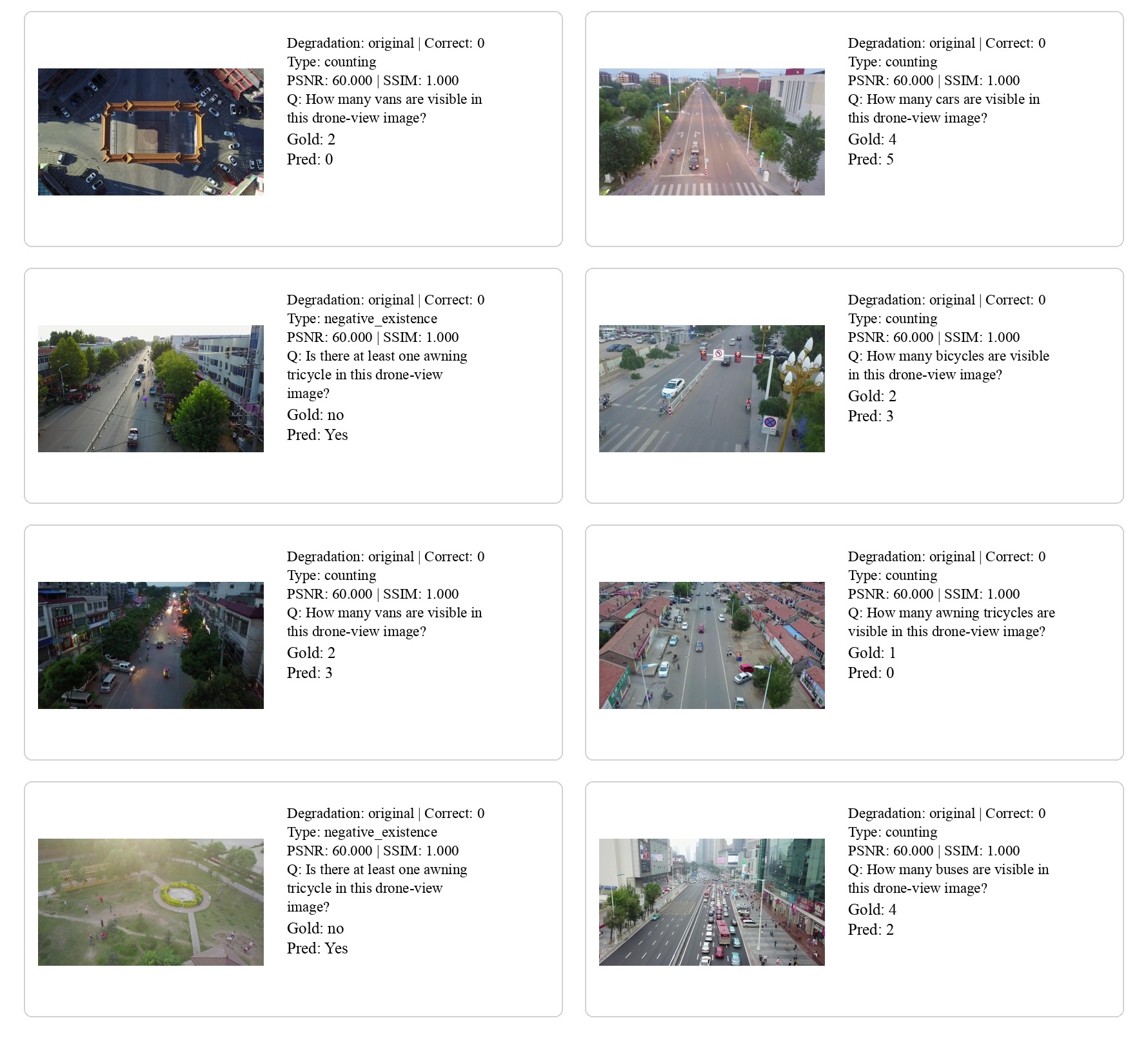}
\caption{Additional VisDrone examples where high perceptual quality still leads to downstream task failure.}
\label{fig:app_visdrone_high_quality_failure}
\end{figure}

\begin{figure}[t]
\centering
\includegraphics[width=\columnwidth]{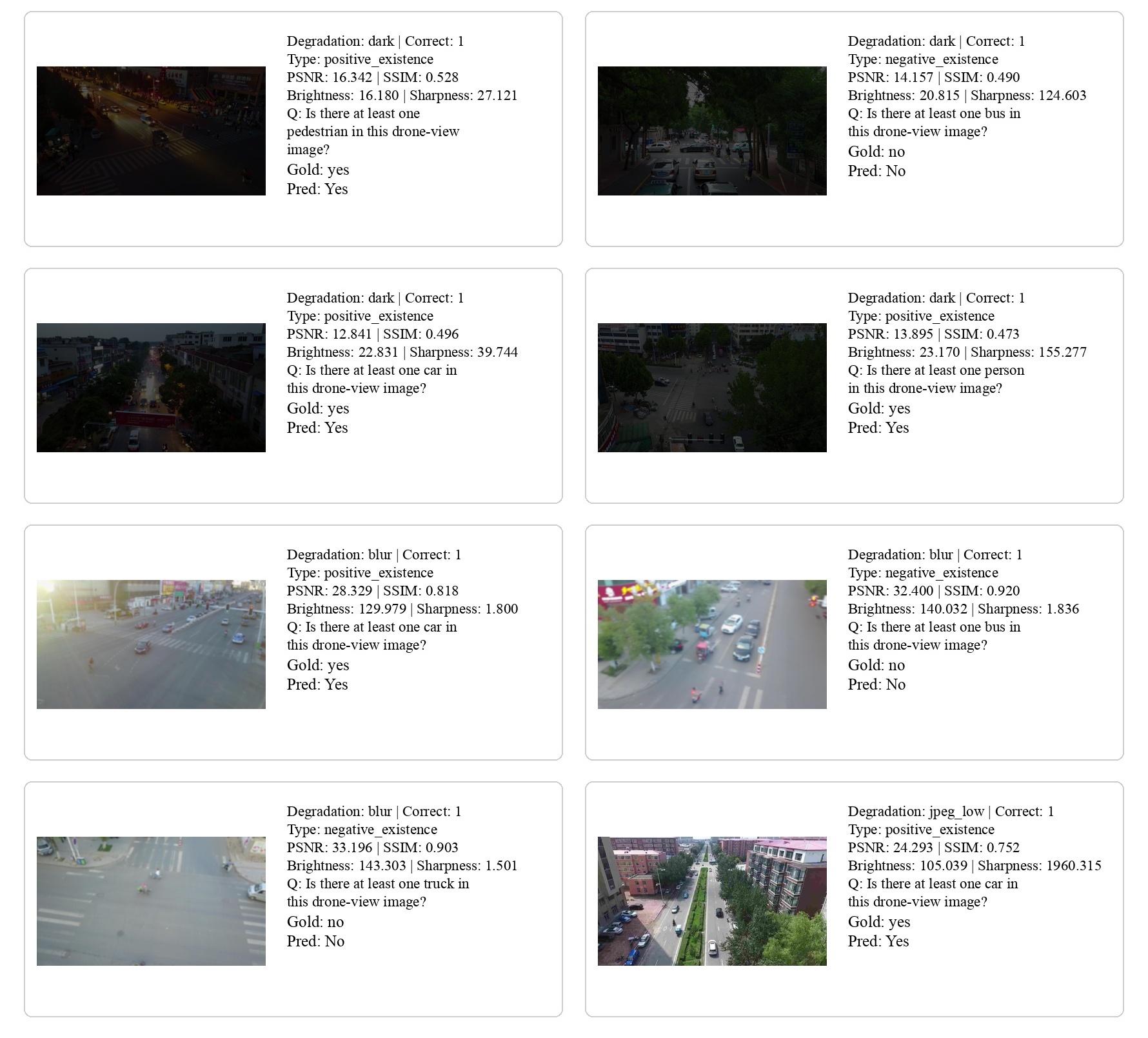}
\caption{Additional VisDrone examples where low perceptual quality still preserves downstream task success.}
\label{fig:app_visdrone_low_quality_success}
\end{figure}

\paragraph{Qualitative cases.}
Figures~\ref{fig:app_visdrone_high_quality_failure} and~\ref{fig:app_visdrone_low_quality_success} illustrate two complementary failure modes of task-agnostic IQA. Some visually clean UAV-view images still fail downstream counting or small-object reasoning tasks, while some degraded images retain sufficient task-relevant cues for correct existence judgment or coarse reasoning. Together, these examples reinforce that embodied IQA should focus on task-relevant information preservation rather than global perceptual quality alone.

\section{Detailed Case Study: UAV Infrastructure Inspection}
\label{app:uav_case_study}

This appendix provides a detailed analysis of how traditional IQA fails in UAV infrastructure inspection and how embodied IQA addresses these limitations.

\subsection{Scenario Description}

A commercial drone inspection system captures images of bridge structures to detect structural damage, particularly cracks and corrosion. The system operates in varying conditions: different lighting conditions, wind speeds, and flight altitudes. Operators need to assess whether captured images are adequate for damage detection before proceeding with analysis.

\subsection{Traditional IQA Failure Modes}

Traditional IQA methods fail in this scenario through several mechanisms. First, they penalize motion blur uniformly, but in bridge inspection, blur in non-critical regions (e.g., sky, background) does not affect crack detection while blur in the target structure does. Second, traditional methods cannot reason about whether critical structures are visible: an image might score highly on general quality metrics but fail to capture the bridge surface needed for inspection. Third, traditional methods provide no explanation, making it difficult for operators to understand why an image is inadequate or how to improve capture parameters.

Prior reports suggest that generic no-reference IQA scores can correlate weakly with inspection outcomes, because they do not encode coverage of the inspected surface or the task-relevant visibility of defects \cite{morgenthal2014quality}. This motivates task-aware assessment criteria that explicitly account for region-level adequacy and semantic coverage.

\subsection{Embodied IQA Approach}

An embodied IQA agent would assess images differently. Given a task prompt such as \emph{assessing whether an image is adequate for detecting structural cracks in the bridge surface}, the agent would identify whether the bridge surface is visible and in focus, evaluate whether motion blur affects critical regions, consider lighting conditions relative to crack visibility requirements, and generate an explanation that links surface visibility, motion blur, and lighting to crack-detection reliability.

This approach addresses the limitations of traditional IQA by providing task-specific reasoning, actionable explanations, and judgments that align with actual inspection needs rather than abstract quality metrics.

\section{Theoretical Extensions}
\label{app:theoretical_extensions}

\subsection{Information-Theoretic Analysis}

From an information-theoretic perspective, task-aware IQA can be understood as maximizing mutual information between quality assessment and task performance \cite{cover1999elements}. Let $H(\text{task\_success}|I)$ denote the entropy of task success given image $I$, and $H(\text{task\_success}|I, Q(I|T,C))$ denote the entropy after conditioning on quality assessment. Task-aware IQA maximizes $I(\text{task\_success}; Q(I|T,C)|I)$, the information that quality assessment provides about task success beyond what is available from the image alone. This information-theoretic framework formalizes why conditional assessment $Q(I|T,C)$ provides more task-relevant information than unconditional assessment $Q(I)$, following the principle that conditional mutual information captures task-specific relevance \cite{cover1999elements}.

Traditional IQA, by contrast, maximizes $I(H(I); Q(I))$, the information shared between human ratings and quality scores. When task-relevant features differ from general quality features, this optimization fails to capture information relevant to task performance, leading to evaluation misalignment \cite{zhang2024embodied}.

\subsection{Generalization Bounds}

We can analyze the generalization properties of embodied IQA through the lens of domain adaptation \cite{ben2010theory}. Traditional IQA methods trained on general quality datasets may not generalize to task-specific scenarios due to domain shift. Embodied IQA, by explicitly conditioning on task and context, reduces this domain shift and improves generalization.

Formally, if we treat different tasks as different domains, embodied IQA performs domain adaptation by conditioning on task specifications. This reduces the effective domain gap compared to methods that assume a single domain (general quality assessment), leading to better generalization across tasks \cite{ganin2016domain}.

\section{Implementation Roadmap}
\label{app:implementation_roadmap}

This appendix outlines a concrete roadmap for implementing embodied IQA systems, organized by component and timeline.

\subsection{Phase 1: Dataset Construction}

The first phase focuses on constructing embodied IQA datasets. This involves collecting images from embodied platforms (drones, robots, AR systems) with associated task specifications, annotating images with task-relative quality judgments and explanations, and measuring actual task performance outcomes. Key challenges include ensuring task diversity, annotation quality, and scale. The required dataset scale depends on task diversity, annotation granularity, and desired generalization; meaningful embodied IQA systems would benefit from diverse task-conditioned examples across multiple domains.

\subsection{Phase 2: Model Development}

The second phase develops MLLM-based IQA models. This involves fine-tuning existing MLLMs (e.g., LLaVA \cite{liu2023visual}, GPT-4V \cite{achiam2023gpt}) on embodied IQA datasets, implementing DPO-based training for task-utility alignment, and developing explanation generation capabilities. Key technical challenges include handling multimodal inputs, generating structured outputs, and ensuring real-time inference. We estimate this phase requires access to large-scale compute resources and expertise in MLLM fine-tuning.

\subsection{Phase 3: Evaluation and Benchmarking}

The third phase establishes evaluation protocols and benchmarks. This involves defining metrics for usefulness (alignment with task performance), explanation fidelity (alignment with visual evidence), and cross-context consistency (appropriate adaptation to changing goals), creating benchmark datasets with ground-truth task performance and expert explanations, and evaluating models on these benchmarks. Key challenges include defining appropriate metrics and ensuring benchmark quality. We estimate this phase requires collaboration with domain experts and extensive validation. See \autoref{sec:eval} in the main text for detailed evaluation principles.

\subsection{Phase 4: Deployment and Integration}

The final phase focuses on real-world deployment. This involves optimizing models for efficiency through distillation and quantization, integrating with embodied systems (UAVs, robots, AR), and validating performance in operational settings. Key challenges include meeting latency requirements, ensuring robustness, and building trust with operators. We estimate this phase requires partnerships with industry and extensive field testing.

This roadmap demonstrates that embodied IQA is not merely a theoretical proposal but a concrete research program with clear milestones and deliverables. The timeline is ambitious but achievable given current MLLM capabilities and the growing availability of embodied datasets.

\section{Practical Risks}
\label{app:practical_risks}
Key risks of the implementation of MLLM-based embodied IQA agents include ungrounded justifications, over-specialization to narrow task prompts, and deployment constraints on latency and compute. Addressing them likely requires joint progress in grounding, efficiency (e.g., distillation \cite{zhang2024embodied}), and benchmark design that rewards safe behavior under uncertainty.

\section{Glossary of Key Concepts}
\label{key_concepts}
To clarify terminology used throughout this paper, we provide definitions of key concepts:

\textbf{Embodied IQA:} Image quality assessment performed from a first-person perspective by an agent that considers task utility, environmental context, and goal-driven reasoning, rather than abstract distortion metrics.

\textbf{Task-aware assessment:} Quality evaluation that considers why an image is useful for a specific goal, reasoning about task relevance, semantic content, and the relationship between visual properties and downstream performance.

\textbf{Conditional quality judgment:} Quality assessment $Q(I \mid T,C)$ that depends on task specification $T$ and contextual information $C$, as opposed to unconditional assessment $Q(I)$ that treats quality as an absolute property.

\textbf{Evaluation misalignment:} A machine learning problem where optimization objectives (e.g., correlation with human ratings) do not match desired outcomes (e.g., task performance), leading to suboptimal system behavior.

\textbf{Task-grounded explanation:} Explanations that relate visual properties to task utility (e.g., ``motion blur obscures the target structure needed for navigation'') rather than merely describing distortions (e.g., ``motion blur in central region'').

%% file: figures/taxonomy_v3.tex
\tikzstyle{leaf}=[draw=hiddendraw,
    rounded corners,minimum height=1 em,
    fill=myblue,text opacity=1, align=center,
    fill opacity=.5,  text=black,align=left,font=\scriptsize,
    inner xsep=2.5 pt,
    inner ysep=1 pt,
    ]
\tikzstyle{middle}=[draw=hiddendraw,
    rounded corners,minimum height=1em,
    fill=output-white!40,text opacity=1, align=center,
    fill opacity=.5,  text=black,align=left,font=\scriptsize,
    inner xsep=3pt,
    inner ysep=1pt,
    ]
\begin{figure*}[ht]
\centering
\begin{forest}
  for tree={
  forked edges,
  grow=east,
  reversed=true,
  anchor=base west,
  parent anchor=east,
  child anchor=west,
  base=middle,
  font=\scriptsize,
  rectangle,
  line width=0.5 pt,
  draw=output-black,
  rounded corners,align=left,
  minimum width=1.5 em,
    s sep=3 pt,
    inner xsep=2 pt,
    inner ysep=1 pt,
  },
  where level=1{text width=4.5 em}{},
  where level=2{text width=6 em,font=\scriptsize}{},
  where level=3{font=\scriptsize}{},
  where level=4{font=\scriptsize}{},
  where level=5{font=\scriptsize}{},
  [\textbf{IQA} with \textbf{MLLMs}, middle, rotate=90, anchor=north, edge=output-black
  [Model Enhancements, middle, text width=6 em
    [Architectural Innovations, middle, text width=7 em
      [CLIP-IQA \cite{wang2023exploring}{,} Q-Instruct \cite{wu2024q}{,} 
      VisualCritic \cite{huang2024visualcritic}{,} \\ PromptIQA \cite{chen2024promptiqa}{,} 
      AgenticIQA \cite{zhu2025agenticiqa}{,} iDETEX \cite{zhao2025idetex}, leaf, text width=26 em]
    ]
    [Reasoning \& Cognitive Methods, middle, text width=9 em
      [Psychophysical Experiments \cite{gao2010image}{,} Dual-Cue Strategy \cite{pan2024multi}{,} \\
      DepictQA* \cite{you2024depicting}{,} Q-Ground* \cite{chen2024q}, leaf, text width=22.5 em]
    ]
  ]
  [Scoring Shift, middle, text width=4em
    [Scoring Paradigm Shift, middle, text width=6.5 em
      [Q-ALIGN  \cite{wu2023q}{,} DeQA-Score \cite{you2025teaching}{,} 
      Teaching LMMs \cite{zhang2025teaching}{,} \\  
      Q-SiT \citep{zhang2025teaching}{,}
      Compare2Score* \cite{zhu2024adaptive}{,} 
      Q-Insight* \cite{li2025q}, leaf, text width=27em]
    ]
    [Evaluation-Driven Optimization, middle, text width=8.5 em
      [AesBench \cite{huang2024aesbench}{,} Q-Bench \cite{zhang2024q}{,} Q-Bench+* \cite{zhang2024q}{,} \\
      Co-Instruct* \cite{wu2024towards}, leaf, text width=26.5em]
    ]
  ]
  [Data-Centric Assessment, middle, text width=7 em
    [Diversity \& Learnability, middle, text width=6.5 em
      [Diversity Entropy \cite{xiao2025data}{,} IQA-Select \cite{li2025exploring}, leaf, text width=18 em]
    ]
   ]
]
\end{forest} 
\caption{A Taxonomy of models for IQA with MLLMs: Model Enhancements, Scoring Shift, and Data-Centric Assessment. Methods marked with * are for comparative multi-image assessment, while others are for single images.}
\label{fig:taxonomy_of_3dllm}
\end{figure*}

%% file: tables/results.tex
\definecolor{lightgray}{gray}{0.92}
\definecolor{highlightgreen}{RGB}{200, 230, 201}
\definecolor{bestgreen}{RGB}{129, 199, 132}
\definecolor{worstred}{RGB}{239, 154, 154}
\definecolor{taskblue}{RGB}{187, 222, 251}
\definecolor{semanticpurple}{RGB}{225, 190, 231}

\begin{table}[H]
\centering
\caption{\textbf{Quantitative Comparison on Task-Oriented IQA.} 
We compare Embodied Agent against Traditional methods (FR/NR) and MLLM-based IQAs, including Q-Insight~\cite{li2025qinsightunderstandingimagequality}, Q-SIT~\cite{zhang2025teachinglmmsimagequality}, and DeQA-Score~\cite{you2025teachinglargelanguagemodels}. 
\textbf{FR}: Full-Reference, \textbf{NR}: No-Reference. 
Note that despite high aesthetic scores, existing MLLM-based IQAs fail to identify safety hazards in some tasks.}
\label{tab:main_result}

\setlength{\aboverulesep}{0pt}
\setlength{\belowrulesep}{0pt}
\renewcommand{\arraystretch}{1.1}

\resizebox{\textwidth}{!}{%
\setlength{\tabcolsep}{2pt}
\begin{tabular}{l ccccc cccc ccc c}
\toprule

\multirow{2}{*}{\textbf{Task Scenario}} &
\multicolumn{5}{c}{\textbf{Full-Reference (FR)}} &
\multicolumn{4}{c}{\textbf{No-Reference (NR)}} &
\multicolumn{3}{c}{\textbf{MLLM-based IQA}} &
\cellcolor{semanticpurple} \\

\cmidrule(lr){2-6}
\cmidrule(lr){7-10}
\cmidrule(lr){11-13}

&
MSE$\downarrow$ & PSNR$\uparrow$ & SSIM$\uparrow$ & LPIPS$\downarrow$ & DISTS$\downarrow$ &
NIQE$\downarrow$ & BRIS.$\downarrow$ & MANI.$\uparrow$ & Hyp.$\uparrow$ &
Q-Ins. & Q-SIT & DeQA. &
\cellcolor{semanticpurple}\multirow{-2}{*}{\raisebox{3.5pt}{\makecell{\textbf{Embodied}\\\textbf{IQA Agent}}}} \\
\midrule

Robotic Grasping &
5925.9 & 10.40 & 0.15 & 0.79 & 0.46 &
8.28 & 23.00 & 0.15 & 0.27 &
3.00 & 1.43 & 1.97 &
\cellcolor{highlightgreen}\textbf{3.50} \\

Aerial Detection &
-- & -- & -- & -- & -- &
3.01 & 1.82 & 0.30 & 0.55 &
3.80 & 4.54 & 2.02 &
\cellcolor{bestgreen}\textbf{5.00} \\

Auto. Driving &
-- & -- & -- & -- & -- &
3.16 & 16.75 & 0.39 & 0.53 &
4.17 & 4.81 & 4.40 &
\cellcolor{worstred}\textbf{1.00}* \\

Authenticity Check &
-- & -- & -- & -- & -- &
3.14 & 24.95 & 0.37 & 0.47 &
3.56 & 4.06 & 3.61 &
\cellcolor{worstred}\textbf{1.00}* \\

Defect Detection &
-- & -- & -- & -- & -- &
2.65 & 14.06 & 0.51 & 0.68 &
3.87 & 4.43 & 4.05 &
\cellcolor{highlightgreen}\textbf{3.00} \\

Indoor Navigation &
-- & -- & -- & -- & -- &
11.87 & 11.87 & 0.30 & 0.34 &
2.30 & 0.59 & 1.83 &
\cellcolor{highlightgreen}\textbf{3.50} \\

\bottomrule

\multicolumn{14}{l}{\footnotesize *Low scores indicate correct detection of critical safety issues (Safety Veto).}
\end{tabular}
}

\vspace{0.5em}

\begin{tcolorbox}[
    colback=teal!5!white,      
    colframe=teal!80!black,    
    arc=2pt,                   
    boxrule=0.8pt,             
    left=4pt, right=4pt, top=4pt, bottom=4pt,
    title=\textbf{Key Insights:}
]
\small 
\begin{itemize}[leftmargin=1.5em, itemsep=1pt, parsep=0pt, topsep=0pt]
    
    \item \textbf{Semantic Safety Veto (Auto. Driving):} While existing MLLMs (Q-SIT: 4.81, DeQA-Score: 4.40) assign high scores based on visual clarity, the Embodied IQA Agent correctly triggers a safety veto (1.00). It prioritizes the semantic hazard ("Wrong-Way Driving") over aesthetic fidelity.

    \item \textbf{Physical Consistency Check (Authenticity Check):} Baselines fail to identify synthetic artifacts (3.61--4.06). In contrast, the Embodied IQA Agent detects physics violations (e.g., floating objects lacking gravity support) and rejects the data as operationally invalid.

    \item \textbf{Task-Grounded Actionability (Indoor Navigation):} Despite low-fidelity sensor noise that severely penalizes traditional metrics (Q-SIT: 0.59), the Embodied IQA Agent assesses the image as actionable (3.50), verifying that the essential traversable geometry remains intact for path planning.

\end{itemize}
\end{tcolorbox}

\end{table}